\documentclass{article} % For LaTeX2e

\usepackage[accepted]{icml2023}
\usepackage{microtype}
\usepackage{graphicx}
\usepackage{subfigure}
\usepackage{booktabs} % for professional tables

\usepackage{url}
\usepackage{hyperref}

\usepackage{dsfont}
\definecolor{dg}{rgb}{0,0.694,0.298}
\definecolor{purple}{rgb}{0.4,0.176,0.569}
\definecolor{royalblue}{RGB}{65,105,225}
\usepackage{pifont}% http://ctan.org/pkg/pifont
\newcommand{\figref}[1]{Fig.~\ref{#1}}
\newcommand{\reqref}[1]{Eq.~\eqref{#1}}
\newcommand{\secref}[1]{Sec.~\ref{#1}}
\newcommand{\tableref}[1]{Table~\ref{#1}}

\makeatletter
\DeclareRobustCommand\onedot{\futurelet\@let@token\@onedot}
\def\@onedot{\ifx\@let@token.\else.\null\fi\xspace}
\def\eg{\emph{e.g}\onedot} 
\def\ie{\emph{i.e}\onedot}

\def\wrt{w.r.t\onedot} 

\makeatother

\definecolor{americanrose}{rgb}{1.0, 0.01, 0.24}

\newcommand{\response}[1]{\textcolor{black}{#1}}

\usepackage{icml2023}

\usepackage[utf8]{inputenc} % allow utf-8 input
\usepackage[T1]{fontenc}    % use 8-bit T1 fonts
\usepackage{hyperref}       % hyperlinks
\usepackage{url}            % simple URL typesetting
\usepackage{booktabs}       % professional-quality tables
\usepackage{amsfonts}       % blackboard math symbols
\usepackage{nicefrac}       % compact symbols for 1/2, etc.
\usepackage{microtype}      % microtypography
\usepackage{xcolor}         % colors
\usepackage{xspace}
\usepackage{amsmath}
\usepackage{graphicx}
\usepackage{enumitem}
\usepackage{subfigure}
\usepackage{adjustbox}
\usepackage{wrapfig}
\usepackage{sidecap}
\usepackage{multirow}
\usepackage{graphicx}

\usepackage[font=scriptsize]{caption}

\hypersetup{
  colorlinks   = true,    % Colours links instead of ugly boxes
  urlcolor     = blue,    % Colour for external hyperlinks
  linkcolor    = blue,    % Colour of internal links
  citecolor    = blue      % Colour of citations
}

\title{FAIRER: Fairness as Decision Rationale Alignment}

\begin{document}

\twocolumn[
\icmltitle{FAIRER: Fairness as Decision Rationale Alignment}
\begin{icmlauthorlist}
\icmlauthor{Tianlin Li}{ntu}
\icmlauthor{Qing Guo}{astar1,astar2}
\icmlauthor{Aishan Liu}{bhu}
\icmlauthor{Mengnan Du}{njit}
\icmlauthor{Zhiming Li}{ntu}
\icmlauthor{Yang Liu}{ztu,ntu}

\end{icmlauthorlist}

\icmlaffiliation{ntu}{Nanyang Technological University, Singapore}
\icmlaffiliation{astar1}{Institute of High Performance Computing (IHPC), Agency for Science, Technology and Research, Singapore}
\icmlaffiliation{astar2}{Centre for Frontier AI Research (CFAR), Agency for Science, Technology and Research, Singapore}
\icmlaffiliation{bhu}{Beihang University, China}
\icmlaffiliation{ztu}{Zhejiang Sci-Tech University, China}
\icmlaffiliation{njit}{New Jersey Institute of Technology, USA}
\icmlcorrespondingauthor{Qing Guo}{tsingqguo@ieee.org}

\icmlkeywords{Machine Learning, ICML}

\vskip 0.3in
]



\printAffiliationsAndNotice{} % otherwise use the standard text.

% \maketitle

\begin{abstract}
Deep neural networks (DNNs) have made significant progress, but often suffer from fairness issues, as deep models typically show distinct accuracy differences among certain subgroups (\eg, males and females).
Existing research addresses this critical issue by employing fairness-aware loss functions to constrain the last-layer outputs and directly regularize DNNs.
Although the fairness of DNNs is improved, it is unclear how the trained network makes a fair prediction, which limits future fairness improvements.
In this paper, we investigate fairness from the perspective of decision rationale and define \response{the \textit{parameter parity score}} to characterize the fair decision process of networks by analyzing neuron influence in various subgroups.
Extensive empirical studies show that the unfair issue could arise from the unaligned decision rationales of subgroups. Existing fairness regularization terms fail to achieve decision rationale alignment because they only constrain last-layer outputs while ignoring intermediate neuron alignment.
To address the issue, we formulate the fairness as a new task, 
\ie, \textit{decision rationale alignment} that requires DNNs' neurons to have consistent responses on subgroups at both intermediate processes and the final prediction.
To make this idea practical during optimization, we relax the naive objective function and propose \textit{gradient-guided parity alignment}, which encourages gradient-weighted consistency of neurons across subgroups.
Extensive experiments on a variety of datasets show that our method can significantly enhance fairness while sustaining a high level of accuracy and outperforming other approaches by a wide margin.
%{https://anonymous.4open.science/r/fairer\_submission-F176/}.
%
\end{abstract}

\section{Introduction}
% 
%-----------------------------------------------------
\begin{figure*}
\centering
\includegraphics[width=0.8\linewidth]{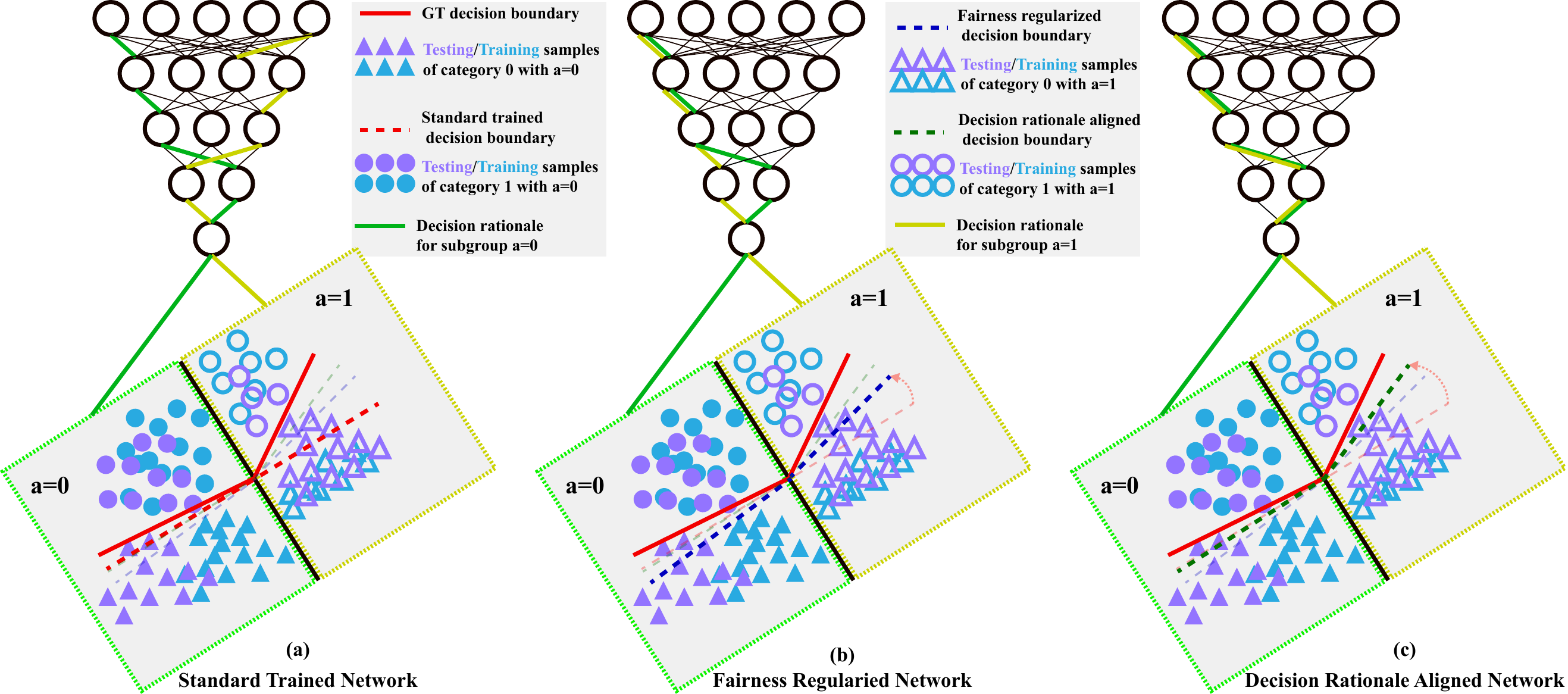}
\caption{Schematic diagrams of two existing solutions and the proposed one. (a) and (b) represent results of the standard trained network and the regularized fairness network. (c) show the results of the decision rationale-aligned network. The previous work, \ie, fairness regularization-based method, adds a regularization term to the final loss function to make the trained network have similar predictions on the two subgroups, which makes the "decision rationales" of the trained network on the two subgroups become partially similar (See the green solid line for the subgroup 
 and yellow solid line for the subgroup 
 in (b)). In contrast, our method is to add a "decision rationale" alignment explicitly and make "decision rationales" on the two subgroups consistent.}
\label{fig:mot}
% \vspace{-15pt}
\end{figure*}
%-----------------------------------------------------

%In the current society, there is a desperate desire for social fairness among individuals. 
%
Deep neural networks (DNNs) are increasingly being used in high-stakes applications in our society.
However, as deep learning is increasingly adopted for many applications that have brought convenience to our daily lives~\citep{resnet,ab,as}, DNNs still suffer from the fairness problem and often exhibit undesirable discrimination behaviors \citep{biasnews, ibmabandon}. 
% For example, for an intelligent task (\eg, face expression classification), a trained DNN easily presents distinct accuracy values in different subgroups (\eg, male and female).
\response{For example, for an intelligent task (\eg, salary prediction), a trained DNN easily presents distinct accuracy values in different subgroups (\eg, males and females).}
The discriminatory behaviors contradict people's growing demand for fairness, which would cause severe social consequences. To alleviate such fairness problems, a line of mitigation strategies has been constantly proposed~\citep{f1,f2,f3}.

A direct regularization method to improve fairness is to relax fairness metrics as constraints in the training process~\citep{Madras2018LearningAF}. This regularization method is designed to reduce the disparities between different subgroups in the training and testing data (See \figref{fig:mot} (a) vs. (b)). 
Although this method easily improves the fairness of DNN models, it is still unclear how the trained network makes a fair decision\footnote{\response{The ‘decision’ here means the prediction results of the DNN regarding given inputs. The name follows the interpretable works \cite{du2019techniques,wang2018interpret,khakzar2021neural}}.}. For example, we do not know \textit{how the fairness regularization terms actually affect the final network parameters and let them make a fair prediction.} 
Without such an understanding, we would not know the effective direction for further fairness enhancement.
Existing work does not address this question and the majority of them concentrate on the last-layer outputs (\ie, predictions) while ignoring the internal process.
In this work, we propose to study the fairness from the perspective of decision rationale and analyze existing fairness-regularized methods through a \textit{decision-rationale-aware analysis} method. 
The term `decision rationale' is known as the reason for making a decision and could be represented as the influence of neurons in a DNN~\citep{khakzar2021neural}.
Specifically, for each intermediate neuron (\ie, a parameter of the DNN \footnote{We follow \citet{molchanov2016pruning,molchanov2019importance} to use the terms "neuron" and "parameter" interchangeably.}), we can calculate the loss change on a subgroup before and after removing the neuron.
As a result, we can characterize the decision rationale of a network on the subgroup by collecting the loss changes of all neurons. For example, the solid green and yellow lines in \figref{fig:mot} represent the neurons leading to high loss changes at each layer and characterize the decision rationales of the two subgroups. 
Then, we define the \textit{parameter parity score} as the decision rationale shifting across different subgroups, which actually reveals the influences of intermediate neurons (\ie, parameters) to the decision rationale changes.
%
% With the new analysis tool, we find that the network fairness is directly related to the consistency of the decision rationales on different subgroups and existing fairness regularization terms could partially achieve this goal (Compare the solid lines in \figref{fig:mot} (b)) since they only add constraints to the final outputs.
\response{
 With the new analysis tool, we find that the network fairness is directly related to the consistency of the decision rationales on different subgroups, and existing fairness regularization terms could only partially achieve this goal, which restricts the fairness improvement (Compare the solid lines in \figref{fig:mot} (b)) since they only add constraints to the final outputs.}
Intuitively, we could define new regularization terms to minimize parity scores of all neurons and encourage them to have similar influence across subgroups.
We name this new task as the \textit{decision rationale alignment} that requires DNNs to have consistent decision rationales as well as final predictions on different subgroups.
Although straightforward, the task is challenging for two reasons: \textit{First}, the decision rationale and parity score are defined based on a dataset and it is impractical to calculate them at each iteration during the training process. \textit{Second}, different neurons have different effects on fairness and such differences should be carefully considered.

%-----------------------------------------------------
% \begin{SCfigure}
% \centering
% \includegraphics[width=0.85\linewidth]{pics/fig_mot.pdf}
% \caption{Schematic diagrams of two existing solutions and the proposed one. (a) and (b) represent results of the standard trained network and the regularized fairness network. (c) show the results of the decision rationale-aligned network.}
% \vspace{-4pt}
% \label{fig:mot}
% \end{SCfigure}

To address the above two challenges, we propose the \textit{gradient-guided parity alignment} method by relaxing the calculation of decision rationale from the dataset-based strategy to the sample-based one. As a result, the corresponding regularization term is compatible with the epoch-based training process. Moreover, we use the first-order Taylor expansion to approximate the parity score between decision rationales, and the effects of different neurons on the fairness are weighted via their gradient magnitudes automatically. Overall, the proposed method can achieve much higher fairness than state-of-the-art methods.
In summary, the work makes the following contributions:
\begin{enumerate}
    \item To understand how a network makes a fair decision, we define \textit{parameter parity score} to characterize the decision rationales of the network on different subgroups. We reveal that the fairness of a network is directly related to the consistency of its decision rationales on different subgroups and existing regularization terms cannot achieve this goal.
    \item To train a fairer network, we formulate the \textit{decision rationale alignment} task and propose the \textit{gradient-guided parity alignment} method to solve it by addressing the complex optimization challenges.
     \item Extensive experiments on three public datasets, \ie, Adult, CelebA, and Credit, demonstrate that our method can enhance the fairness of DNNs effectively and outperform others largely.

\end{enumerate}

% \input{sections/preliminaries}

% -----------------------------------------------------
% \input{sections/methodology}

\section{Preliminaries} %: Group Fairness via Regularization Losses

% In this section, we first formulate the fairness task and introduce related notations. Then, we present existing methods that add fairness regularization terms in the loss functions to train fair DNNs. Finally, we show that such methods have remaining unfair issues with some accuracy sacrifices. 

\subsection{Problem Formulation}

In general, given a dataset $\mathcal{D}$ containing data samples (\ie, $\mathbf{x}\in\mathcal{X}$) and corresponding labels (\ie, $y\in \mathcal{Y}$), we can train a DNN to predict the labels of input samples, \ie, $\hat{y}=\text{F}(\mathbf{x})$ with $\hat{y}\in \mathcal{Y}$ being the prediction results. 
In the real world, the samples might be divided into subgroups according to some sensitive attributes $a\in\mathcal{A}$ such as gender and race. Without loss of generality, we consider the binary classification and binary attribute setup, \ie, $y\in\{0,1\}$ and $a\in\{0,1\}$. For example, $a=0$ and $a=1$ could represent males and females, respectively.
%
% When we have a testing dataset $\mathcal{T}$ that contains two subsets $\mathcal{T}_{a=0}$ and $\mathcal{T}_{a=1}$,
%
A fair DNN (\ie, $\text{F}(\cdot)$) is desired to obtain a similar accuracy in the two subgroups. 

\subsection{Fairness Regularization}
\label{sec:fairess}
% Existing fairness works \citep{dp,eo,Madras2018LearningAF,fairmixup} focus on designing fairness regularization terms and adding them to the loss function, which encourages the targeted DNN to predict similar results across subgroups. Specifically, \citet{dp} develop the demographic parity (DP) regularization term to encourage the predicted label to be independent of the sensitive attribute (\ie, $a$), that is, $P(\hat{y}|a=0)=P(\hat{y}|a=1)$ which means that the probability distribution of $\hat{y}$ condition on $a=0$ should be the same as the condition on $a=1$. \citet{eo} further propose the equalized odds (EO) regularization to consider the ground truth label $y$ and make the prediction and sensitive attribute conditionally independent \wrt $y$, \ie, $P(\hat{y}|a=0, y)=P(\hat{y}|a=1,y)$. Although straightforward, it is difficult to optimize the above regularization terms and \citet{Madras2018LearningAF} propose relaxed counterparts:
\response{ Among fairness evaluation measures, Demographic Parity (DP) and Equalized Odds (EO) are most frequently adopted in deep learning fairness research. Specifically, \citet{dp} develop the DP metric to encourage the predicted label to be independent of the sensitive attribute (\ie, $a$), that is, $P(\hat{y}|a=0)=P(\hat{y}|a=1)$ which means that the probability distribution of $\hat{y}$ condition on $a=0$ should be the same as the condition on $a=1$. \citet{eo} further propose the EO metric to consider the ground truth label $y$ and make the prediction and sensitive attribute conditionally independent \wrt $y$, \ie, $P(\hat{y}|a=0, y)=P(\hat{y}|a=1,y)$. Although straightforward, it is difficult to optimize the above measures and existing fairness works \citep{Madras2018LearningAF,fairmixup} focus on designing fairness regularization terms and adding them to the loss function, which encourages the targeted DNN to predict similar results across subgroups. \citet{Madras2018LearningAF} propose relaxed counterparts: }
%
%-----------------------
\begin{align} \label{eq:DP}
\Delta \text{DP}(\text{F}) = \left| \text{E}_{\mathbf{x} \sim P_0}(\text{F}(\mathbf{x})) - \text{E}_{\mathbf{x} \sim P_1}(\text{F}(\mathbf{x}))\right|,
\end{align}
%-----------------------
where $P_0=P(\mathbf{x}|a=0)$ and $P_1=P(\mathbf{x}|a=1)$ are the distributions of $\mathbf{x}$ condition on $a=0$ and $a=1$, respectively, and the function $\text{E}(\cdot)$ is to calculate the expectation under the distributions. 
%
%-----------------------
\begin{align} \label{eq:EO}
\Delta \text{EO}(\text{F}) = \sum_{y \in \{0, 1\}}\left| \text{E}_{\mathbf{x} \sim P_0^y}(\text{F}(\mathbf{x})) - \text{E}_{\mathbf{x} \sim P_1^y}(\text{F}(\mathbf{x}))\right|,
\end{align}
%-----------------------
%
where $P_0^1=P(\mathbf{x}|a=0,y=1)$ denotes the distribution of $\mathbf{x}$ condition on the $a=0$ and $y=1$, and we have similar notations for $P_0^0$, $P_1^1$, $P_1^0$ if we set the DNN for a binary classification task and have the label $y\in{0,1}$.  
We can add \reqref{eq:DP} and \reqref{eq:EO} to the classification loss (\eg, cross-entropy loss) to regularize the fairness of the targeted DNN, respectively, and obtain the whole loss function
%
%-----------------------
\begin{align} \label{eq:LossDPorEP}
\mathcal{L}= \text{E}_{(\mathbf{x},y) \sim P}(\mathcal{L}_{\text{cls}}(\text{F}(\mathbf{x}),y)) + \lambda \mathcal{L}_\text{fair}(\text{F}),
\end{align}
%-----------------------
%
where $P$ denotes the joint distribution of $\mathbf{x}$ and $y$, $\mathcal{L}_{\text{cls}}$ is the classification loss, and the term $\mathcal{L}_\text{fair}$ could be $\Delta \text{DP}(\text{F})$ or $\Delta \text{EO}(\text{F})$ defined in \reqref{eq:DP} or \reqref{eq:EO}. 
We can minimize the above loss function and get fairness-regularized DNNs.
Although effective, the above method presents some generalization limitations. To alleviate this issue, \cite{fairmixup} embed the data augmentation strategy into the fairness regularization method and propose FairMixup with novel DP- and EP-dependent regularization terms. Please refer to \citet{fairmixup} for details. 
% \response{Our method is also applicable to other fairness metrics that quantify the expected difference between groups. More analysis is put in supplementary materials.}

Overall, we get several fairness regularization methods via different regularization terms. Specifically, we denote the methods without augmentation as $\text{FairReg}(\Delta \text{DP}, \text{noAug})$ and $\text{FairReg}(\Delta \text{EO}, \text{noAug})$ based on regularization functions (\ie, \reqref{eq:DP} and \reqref{eq:EO}).
We denote the methods equipped with data augmentation as $\text{FairReg}(\Delta \text{DP}, \text{Aug})$ and $\text{FairReg}(\Delta \text{EO}, \text{Aug})$, respectively.

% \subsection{Limitations }
\subsection{\response{Observations}}
\label{subsec:limitation}

%-----------------------------------------------------
% \begin{wrapfigure}{R}{0.45\textwidth}
% \centering
% \includegraphics[width=1\linewidth]{pics/fig_adult_DP.pdf}
% \vspace{-10pt}
% \caption{Accuracy and fairness comparison of five different methods on the Adult dataset. The hyperparameter $\lambda$ increases from 0.2 to 0.6 along the $-\text{DP}$ axis as it becomes larger.} 
% % \litl{data changed.}}
% \label{fig:adult_DP}
% \vspace{-10pt}
% \end{wrapfigure}

\begin{figure}
\centering
\includegraphics[width=1\linewidth]{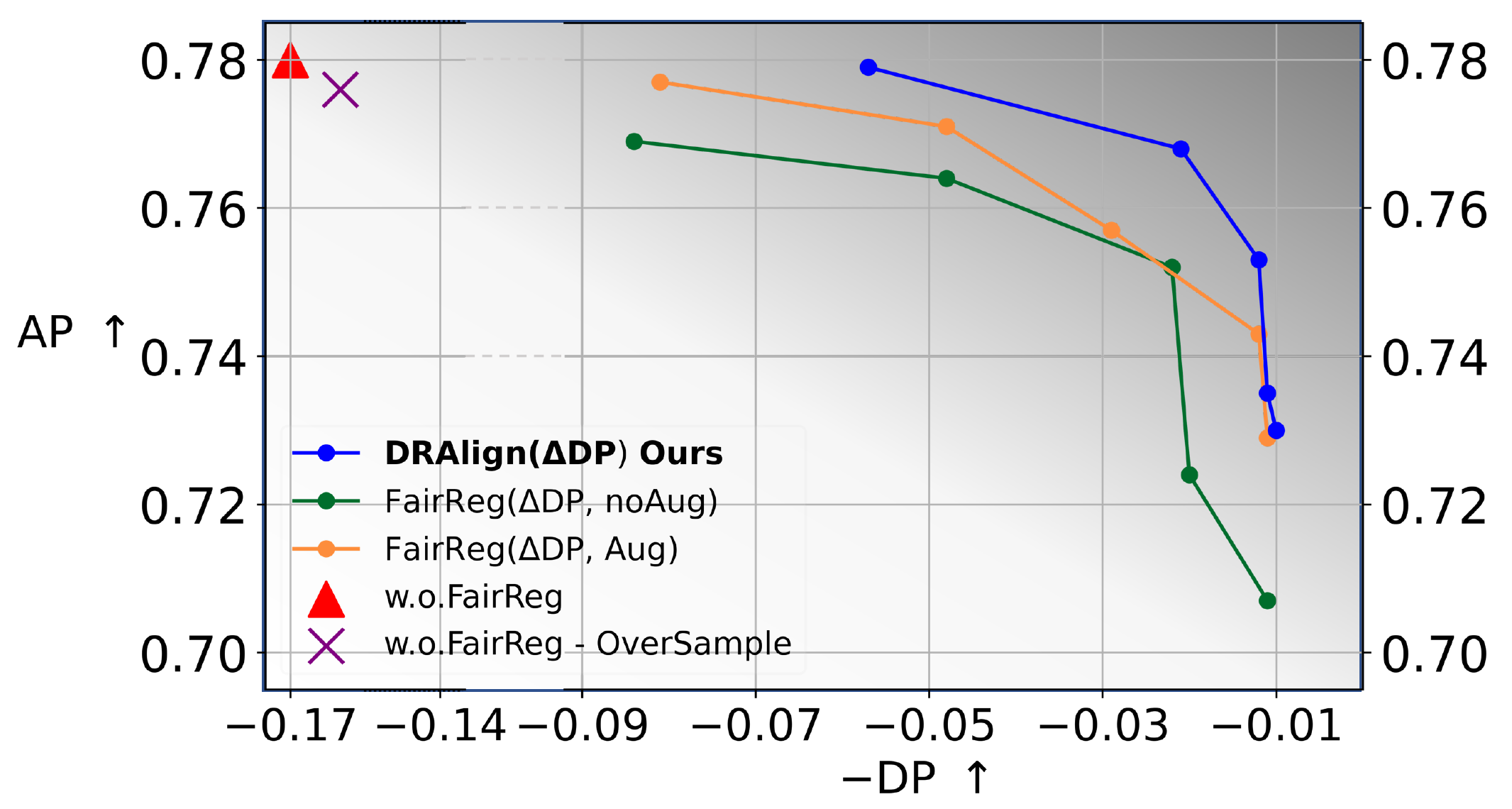}
\caption{Accuracy and fairness comparison of five different methods on the Adult dataset. The hyperparameter $\lambda$ increases from 0.2 to 0.6 along the $-\text{DP}$ axis as it becomes larger.} 
% \litl{data changed.}}
\label{fig:adult_DP}
% \vspace{-15pt}
\end{figure}
%-----------------------------------------------------

% Although the above methods are able to enhance the fairness of DNNs, they still present some limitations.
%
We conduct an experiment on the Adult dataset \citep{adult} with a neural network with 3-layer MLPs. 
Specifically, we train the network with two fairness regularization methods (\ie, $\text{FairReg}(\Delta \text{DP}, \text{noAug})$ and $\text{FairReg}(\Delta \text{DP}, \text{Aug})$ \footnote{We have similar observations on the $\Delta \text{EO}$-based methods and remove them for a clear explanation.}) and five different $\lambda\in\{0.2, 0.3, 0.4, 0.5, 0.6\}$, that is, for each method, we get five trained networks. 
Then, we can calculate the accuracy scores and fairness scores of all networks on the testing dataset. We employ average precision for the accuracy score and $-\text{DP}$ for the fairness score since a smaller $\text{DP}$ means better fairness.
For each method, we can draw a plot \wrt different $\lambda$.
Besides, we also train a network without the fairness regularization term and denote it as $\text{w.o.FairReg}$. Based on $\text{w.o.FairReg}$, we can conduct oversampling on the training samples to balance the samples across different subgroups \citep{stra} and denote it as $\text{w.o.FairReg-Oversample.}$
As shown in Fig.\ref{fig:adult_DP}, we see that: \ding{182} The standard trained network via $\text{w.o.FairReg}$ presents an obvious fairness issue and the oversampling solution has limited capability to fix it.
\ding{183} \response{When we use the regularization methods and gradually increase the weight $\lambda$ in \reqref{eq:LossDPorEP} from 0.2 to 0.6, $\text{FairReg}(\Delta \text{DP}, \text{noAug})$ is able to generate fairer networks with higher fairness scores (\ie, higher -$\text{DP}$) than the one from $\text{w.o.FairReg}$. However, the corresponding accuracy decreases by a large margin, that is, \textit{existing methods could hardly generate enough fair networks under similar accuracy.}}
% For example, when we set $\lambda$ as 0.5, the $\Delta \text{DP}$ value decreases from around 0.16 to about 0.04 with $75$\% relative changes, which means a significant fairness enhancement. Meanwhile, the accuracy decreases from around 0.78 to 0.72. 
%
\ding{184} The data augmentation-based method (\ie, $\text{FairReg}(\Delta \text{DP}, \text{Aug})$) can alleviate such an issue to some extent and achieves higher fairness than $\text{FairReg}(\Delta \text{DP}, \text{noAug})$ under similar accuracy. 
%

% The above observations motivate us to study the reason behind this and how to further address this issue. Intuitively, a straightforward reason is that the optimization directions of the network parameters driven by fairness regularization and classification loss are different. However, the influence on trained neural networks is unknown and should be carefully studied.
\response{Such fairness regularization methods neglect the decision-making process and might generate sub-optimal models. Although intuitively having a consistent decision process among various groups could enhance model performance in terms of fairness, we still empirically explore the connection between the decision-making process and fairness. We provide an analysis method by extending the decision rationale-aware explainable methods in Sec. \ref{sec:neuron_dis_analysis}. Specifically, instead of using the final fairness metrics, we define the parameter parity score for each parameter of a network that measures whether the parameter is fair, that is, whether it has consistent responses to different subgroups.}

\section{Decision Rationale-aware Fairness Analysis}
\label{sec:neuron_dis_analysis}

In recent years, decision rationale-aware explainable methods are developed and help understand how a trained network makes a decision \citep{khakzar2021neural,wang2018cvpr}. In these works, the decision rationale is represented by measuring the importance of intermediate neurons.
Inspired by this idea, to understand a fair decision, we study the decision process of networks by analyzing their neuron influence under different subgroups, and define the decision rationales for different subgroups.
Then, we define the \textit{parity score} for a network that actually measures whether the decision rationales on different subgroups are consistent. Besides, we can use the parity score to compare the networks trained with different regularization terms.

\subsection{Parameter Parity Score}
\label{subsec:neuron_dis_weight}

Inspired by recent work on understanding the importance of the neuron for the classification loss~\citep{molchanov2019importance}, we define the parameter parity score based on the independent assumption across neurons (\ie, parameters)\footnote{More details about this assumption are deferred to \ref{sec:assumptions}.}.
%
% Given a sample dataset $\mathcal{D}$, we can split it to two subsets (\ie, $\mathcal{D}_{a=0}$ and $\mathcal{D}_{a=1}$) corresponding to different subgroups defined by a sensitive attribute. 
%
When we have a trained network $\text{F}(\cdot)$ with its parameters $\mathcal{W} =\{w_0,\ldots,w_K\} $, we can calculate classification losses on samples from two distributions $P_0=P((\mathbf{x},y)|a=0)$ and $P_1=P((\mathbf{x},y)|a=1)$ \response{which correspond to the training subsets of two subgroups (\ie, $a=0$ and $a=1$), and get the losses $\mathcal{J}(\text{F},P_0)$ and $\mathcal{J}(\text{F},P_1)$, respectively. }
%
% $\mathcal{D}_{a=0}$ and $\mathcal{D}_{a=1}$, which are denoted as $\mathcal{J}(\text{F},\mathcal{D}_{a=0})$ and $\mathcal{J}(\text{F},\mathcal{D}_{a=1})$, respectively. 
%
Meanwhile, we can modify $\text{F}(\cdot)$ by removing a specific parameter $w_k$ and denote the new counterpart as $\text{F}_{w_k=0}$, and we can also obtain losses via $\mathcal{J}(\text{F}_{w_k=0},{P}_{0})$ and $\mathcal{J}(\text{F}_{w_k=0},{P}_{1})$.
Then, for each subgroup (\ie, $P_0$ or $P_1$), we calculate the loss change before and after removing the parameter $w_k$ by
%
%-----------------------
\begin{align} \label{eq:lossdiff}
c_k^{a=i}=\text{C}(\text{F}, w_k, P_i)
= |\mathcal{J}(\text{F},P_i) - \mathcal{J}(\text{F}_{w_k=0},P_i)|^2, \\
\nonumber 
\forall i\in\{0,1\},k\in[0,K],
\end{align}
%-----------------------
%
where the function $\mathcal{J}(\text{F},P_i)$ is to calculate the classification loss (\ie, $\mathcal{L}_\text{cls}$ in \reqref{eq:LossDPorEP}) of examples in $P_i$ with $\forall i\in\{0,1\}$ based on the network $\text{F}$. 
With a subgroup $P_i$ and a $K$-neuron network $\text{F}$, we can get $\mathbf{c}_\text{F}^{a=i}=[c_0^{a=i},c_1^{a=i},\ldots, c_K^{a=i}]$ that is regarded as a representation of the decision rationale on the subgroup $P_i$  \citep{khakzar2021neural}.

Then, we define the parity score of the parameter $w_k$ as the difference between $c_k^{a=0}=\text{C}(\text{F}, w_k, P_0)$ and $c_k^{a=1}=\text{C}(\text{F}, w_k, P_1)$, \ie, 
%
%-----------------------
\begin{align} \label{eq:dis_parameter}
d_k = |\text{C}(\text{F}, w_k, P_0)-\text{C}(\text{F}, w_k, P_1)|^2.
\end{align}
%-----------------------
%
Intuitively, if the network $\text{F}$ is fair to a kind of sensitive attribute, each parameter should have consistent responses to different subgroups, and the changes before and after removing the parameter should be the same. As a result, a smaller $d_k$ means that the parameter $w_k$ is less sensitive to the attribute changes.
For the entire network with $K$ neurons, we get $K$ parity scores and $\mathbf{c}_\text{F}^{a=i} = [c_0^{a=i},c_1^{a=i},\ldots,c_K^{a=i}]$, and can represent the network with $\mathbf{d}_{\text{F}}=[d_0,d_1,\ldots,d_K]$ and aggregate all scores for a network-level parity score, \ie, $d_{\text{F}}=\sum_{k=0}^Kd_k=|\mathbf{c}_\text{F}^{a=0}-\mathbf{c}_\text{F}^{a=1}|_1$, which measures whether the decision rationales on the two subsets are consistent (\ie, properly aligned).  
%
% \litl{form a layer-wise vector composed by ck. Then normalize this vector first. }

\subsection{Relationship between Parity Score and Fairness}
\label{sec:neuron_parity_score}
With the \textit{parameter parity score}, we conduct an empirical study based on the Adult dataset and a neural network with 3-layer MLPs. Specifically, we train six networks with the regularization terms defined in \secref{subsec:limitation}, \eg, the $\Delta \text{DP}$-based regularization terms with six different weights (\ie, $\text{FairReg}(\Delta \text{DP}, \text{noAug})$ with $\lambda\in\{0.0, 0.2,0.3,0.4,0.5,0.6\}$). Note that, $\text{FairReg}(\Delta \text{DP}, \text{noAug})$ with $\lambda=0.0$ represents the standard trained network without fair regularization terms (\ie, $\text{w.o.FairReg}$).
Then, for each method, we can train a neural network and calculate the parity score, \ie, $d_\text{F}=\sum_{k=0}^Kd_k=|\mathbf{c}_\text{F}^{a=0}-\mathbf{c}_\text{F}^{a=1}|_1$ to measure the decision rationale shifting across subgroups and the fairness score defined by $-\text{DP}$. 
As reported in \tableref{tab:tab1}, we see that: \ding{182} the parity score of the network gradually decreases as the $\text{DP}$ becomes smaller, which demonstrates that the fairness of a network is highly related to the decision rationale shifting across subgroups.
\ding{183} adding the fairness regularization term on the last-layer outputs (\ie, $\lambda>0$) can decrease the decision rationale shifting to some extent. However, such an indirect way could hardly achieve the optimized results and a more effective way is to actively align the decision rationale explicitly.
Note that we can observe similar results on other regularization methods and focus on $\text{FairReg}(\Delta \text{DP}, \text{noAug})$ due to the limited space. 
\response{We conclude that the existing fairness regularization-based methods can only encourage the consistency between decision rationales of the network on different subgroups to some extent.}
\response{This inspires our method in Sec.\ref{sec:DRAchapter} that conducts alignment of the decision rationales of different subgroups explicitly.}
\response{It is worth noting that the parameter parity score is the most straightforward way to measure whether the parameter has consistent responses to different subgroups and represents the degree of the decision rationale alignment.}
% , and is inspired by the decision process-related works that analyze the neuron behaviors under different subgroups.}

% \litl{no fair mixup. fair mixup does not work well on other settings. More similar experiments are put to Section 6.3.1}

\begin{table}[t]
    \centering
    \caption{Parity scores, fairness scores, and the first-order Taylor approximation of the parity scores of networks trained via $\text{FairReg}(\Delta \text{DP}, \text{noAug})$ with different $\lambda$ in \reqref{eq:LossDPorEP}. For each network,  we train 10 runs with different seeds and the average results are reported.} 
    % \begin{adjustbox}{max width=\textwidth,keepaspectratio}
    % \resizebox{0.85\columnwidth}{!}{%
    \resizebox{0.95\columnwidth}{!}{%
    \begin{tabular}{c|c|c|c|c|c|c}
    \toprule
         &  \multicolumn{6}{c}{$\text{FairReg}(\Delta \text{DP}, \text{noAug})$} \\
          &            $\lambda=0.0$ & $\lambda=0.2$ & $\lambda=0.3$ & $\lambda=0.4$  & $\lambda=0.5$ & $\lambda=0.6$ \\ \midrule 
        % Decision rationale shifting ($d_{\text{F}}$) & $0.6236$  & $0.391 \pm 0.338$ & $0.101 \pm 0.084$ & $0,070 \pm 0.045$  & $0.0458 \pm 0.023$   & $0.039 \pm 0.029$ \\ \midrule
        Parity score ($d_{\text{F}}$) & $0.624$  & $0.391$ & $0.101$ & $0.070$  & $0.046$   & $0.039$ \\ \midrule
        Fairness ($-\text{DP}$) & $-0.160$ & $-0.084$ & $-0.048$ & $-0.022$ & $-0.020$ & $-0.010 $ \\  \midrule
        Approx. ($-\sum_{l=0}^L \text{cos}(\vec{\mathbf{c}}_l^{a=0},\vec{\mathbf{c}}_l^{a=1})$) & $-0.670$  & $-1.382$ & $-1.530$ &  $-1.629$ & $-1.631$ & $-1.800$ \\
         \bottomrule 
    \end{tabular}
    }
    % \end{adjustbox}
\label{tab:tab1}
\end{table}

\section{Decision Rationale Alignment}
\label{sec:DRAchapter}
\subsection{Formulation and Challenges}

According to \reqref{eq:dis_parameter}, we can achieve a fairer network by aligning the decision rationales of subgroups and a straightforward way is to set the parity score $d_{\text{F}}=\sum_{k=0}^Kd_k$ as an extra loss function and minimize it directly, that is, we can add a new loss to \reqref{eq:LossDPorEP}  and have 
%
%-----------------------
\begin{align} \label{eq:LossDRA}
\mathcal{L}= \text{E}_{(\mathbf{x},y) \sim P}(\mathcal{L}_{\text{cls}}(\text{F}(\mathbf{x}),y)) + \lambda \mathcal{L}_\text{fair}(\text{F}) + \beta \sum_{k=0}^Kd_k,
\end{align}
%-----------------------
%
where $d_k$ is the parity score of the $k$th neuron and calculated by \reqref{eq:dis_parameter}. Such a loss should calculate parity scores for all neurons and all samples in a dataset, leading to a high cost and is not practical.

\subsection{Gradient-guided Parity Alignment}
\label{subsec:gradient-guided}

To address the challenges, we relax \reqref{eq:lossdiff} to the sample-based counterpart
%
%-----------------------
\begin{align} \label{eq:lossdiff_dist}
c_k^{a=i}= \text{C}(\text{F}, w_k, P_i)
= & |\text{E}_{(\mathbf{x},y)\sim P_i}(\mathcal{L}_\text{cls}(\text{F}(\mathbf{x}),y) - \\
\nonumber 
& \text{E}_{(\mathbf{x},y)\sim P_i}(\mathcal{L}_\text{cls}(\text{F}_{w_k=0}(\mathbf{x}),y)) |^2, \\ \nonumber 
& \forall i\in\{0,1\},k\in[0,K].
\end{align}
%-----------------------
%
We use the first-order Taylor expansion to approximate $c_k^{a=i}$ similar to \citet{molchanov2019importance} and get
%
%-----------------------
\begin{equation} 
\small
\label{eq:lossdiff_dist_approx}
\hat{c}_k^{a=i}=\hat{\text{C}}(\text{F}, w_k, P_i)
=  (g_k^{a=i}\cdot w_k)^2,~\forall i\in\{0,1\},k\in[0,K].
\end{equation}
%-----------------------
%
where $g_k^{a=i}$ denotes the gradient of the $k$th neuron (\ie, $w_k$) \wrt the loss function on the examples sampled from the distribution of the $i$th subgroup (\ie, $P_i$).
Intuitively, the above definition means that we should pay more attention to the neurons with higher gradients and make them have similar responses to different subgroups. 
However, neurons (\ie, parameters) of different layers may have different score ranges. To avoid this influence, we further normalize $\hat{c}_k^{a=i}$ by $\frac{\hat{c}_k^{a=i}} {|\hat{\mathbf{c}}_l^{a=i}|}~\forall i\in\{0,1\}, k\in \mathcal{K}_l$, where $\mathcal{K}_l$ contains the indexes of the neurons in the $l$th layer, and parity scores of neurons in the $l$th layer (\ie, $\{\hat{c}_k^{a=i}|k\in \mathcal{K}_l\}$) form a vector $\hat{\mathbf{c}}_l^{a=i}=\text{vec}(\{\hat{c}_k^{a=i}|k\in \mathcal{K}_l\})$.
Then, we can get a new vector for the $l$th layer $\vec{\mathbf{c}}_l^{a=i}=\text{vec}(\{\frac{\hat{c}_k^{a=i}} {|\hat{\mathbf{c}}_l^{a=i}|}|k\in \mathcal{K}_l\}$) by normalizing each element.
% %
% \begin{align}  \label{eq:lossdiff_dist_approx_norm}
% \vec{c}_k^{a=i} = \frac{\hat{c}_k^{a=i}} {|\hat{\mathbf{c}}_l^{a=i}|},~\forall i\in\{0,1\}, k\in \mathcal{K}_l, \hat{\mathbf{c}}_l^{a=i}=\text{vec}(\{\hat{c}_k^{a=i}|k\in \mathcal{K}_l\})
% \end{align}
% %
% where $\mathcal{K}_l$ contains the indexes of the neurons (\ie, parameters) in the $l$th layer, and parity scores of neurons in the layer (\ie, $\{\hat{c}_k^{a=i}|k\in \mathcal{K}_l\}$) form a vector $\hat{\mathbf{c}}_l^{a=i}$. 
%
Then, we can update \reqref{eq:LossDRA} by minimizing the distance between $\vec{\mathbf{c}}_l^{a=0}$ and $\vec{\mathbf{c}}_l^{a=1}$ $\forall l\in [0, L]$, \ie,
%
%-----------------------
\begin{align} \label{eq:LossDRA_update}
\mathcal{L} & = \text{E}_{(\mathbf{x},y) \sim P}(\mathcal{L}_{\text{cls}}(\text{F}(\mathbf{x}),y)) + \lambda \mathcal{L}_\text{fair}(\text{F}) \\
\nonumber
& - \beta \sum_{l=0}^L \text{cos}(\vec{\mathbf{c}}_l^{a=0}, \vec{\mathbf{c}}_l^{a=1}),
\end{align}
%-----------------------
%
where $L$ denotes the number of layers in the network, and the function $\text{cos}(\cdot)$ is the cosine similarity function. The last two terms are used to align the final predictions and the responses of the intermediate neurons across subgroups, respectively. 
To validate the approximation (\ie, $-\sum_{l=0}^L \text{cos}(\vec{\mathbf{c}}_l^{a=0}, \vec{\mathbf{c}}_l^{a=1})$) can reflect the decision rationale alignment degree like the parity score $\sum_{k=0}^Kd_k$, we conduct an empirical study on $\text{FairReg}(\Delta \text{DP}, \text{noAug})$ as done in \secref{sec:neuron_parity_score} and calculate the value of $-\sum_{l=0}^L \text{cos}(\vec{\mathbf{c}}_l^{a=0}, \vec{\mathbf{c}}_l^{a=1})$ for all trained networks. From \tableref{tab:tab1}, we see that the approximation has a consistent variation trend with the parity score under different $\lambda$.

% \litl{Moreover, the approximation enables us to calculate the $\text{cos}(\vec{\mathbf{c}}_l^{a=0}, \vec{\mathbf{c}}_l^{a=1})$ to evaluate the decision rationale alignment degree with lower computation cost than \textit{parameter parity score}.} 
% \litl{Similar to the experiments in section \ref{sec:neuron_parity_score}, we conduct an empirical study to show that the approximation could still reflect the decision rationale alignment degree. We calculate the decision alignment degree based on the Adult dataset and 3-layer MLP models.}

% From the Table ~\ref{tab:cosine}, we can see that the alignment degree improves with increasing $\lambda$. Our approximation to the \textit{parameter parity score} could measure whether the decision rationales on different subgroups.

% % \subsection{\litl{Validity of the approximation. Correlation}}
% \begin{table}[!ht]
%     \centering
%     \caption{Analysis for correlation between the cosine similarity and fairness regularization. The stricter the fairness regularization, the higher the decision rationale alignment degree.} 
%     \begin{adjustbox}{max width=\textwidth}

%     \begin{tabular}{c|c|c|c|c|c|c}
%     \toprule
%           & Standard Training & $\lambda=0.2$ & $\lambda=0.3$ & $\lambda=0.4$ & $\lambda=0.5$ & $\lambda=0.6$\\ \midrule 
%         $\sum_{l=0}^L \text{cos}(\vec{\mathbf{c}}_l^{a=0}, \vec{\mathbf{c}}_l^{a=1})$ & 0.67  & 1.38 & 1.53 &  1.63 & 1.63 & 1.80 \\ \bottomrule
%     \end{tabular}
%     \end{adjustbox}
% \vspace{-0.1in}
% \label{tab:cosine}
% \end{table}

\subsection{Implementation Details}
\begin{algorithm}[t]

% \small
%   \LinesNumberedHidden
\caption{Gradient-guided Parity Alignment}
\label{alg:alg}

    \begin{algorithmic}[1]
    
    \STATE {\bfseries Input:} Network $\text{F}$ with parameters $\mathcal{W}=\{w_0,\ldots,w_K\}$, epoch index set $\mathcal{E}$, training data $\mathcal{D}$, 
    % % iteration index set $\mathcal{I}$ per epoch,
    batch size $B$, network layers $L$, neurons in the $l$th layer $\mathcal{K}_l$, hyper-parameters $\lambda$ and $\beta$, learning rate $\eta$
    % \REPEAT
    % \STATE Initialize $noChange = true$.
    \FOR{$e \in \mathcal{E}$}
    \STATE // Sampling $B$ samples from subgroups in $\mathcal{D}$
    \STATE $(\mathbf{X}_0, \mathbf{Y}_0) \leftarrow \text{Sample}(\mathcal{D}, a=0, B)$
    \STATE $(\mathbf{X}_1, \mathbf{Y}_1) \leftarrow \text{Sample}(\mathcal{D}, a=1, B)$
    \STATE // Calculating loss and updating the model
    \STATE $\mathcal{L}_{\text{cls}} = \mathcal{L}_{\text{cls}}(\text{F}(\mathbf{X}_0),\mathbf{Y}_0)$ + $\mathcal{L}_{\text{cls}}(\text{F}(\mathbf{X}_1),\mathbf{Y}_1)$
    \STATE $\mathcal{L}_\text{fair} =\Delta \text{DP}(\mathbf{F},\mathbf{X}_0,\mathbf{X}_1)$
    \FOR{$l \in \mathcal{L}$}
    \FOR{$k \in \mathcal{K}_l$}
    \STATE $g_k^{a=0} =  \frac{\partial(\mathcal{L}_{\text{cls}}(\text{F}(\mathbf{X_0}),\mathbf{Y_0}))}{\partial w_k}$
    \STATE $g_k^{a=1} = \frac{\partial(\mathcal{L}_{\text{cls}}(\text{F}(\mathbf{X_1}),\mathbf{Y_1}))}{\partial w_k}$ 
    \STATE $\hat{c}_k^{a=0} = (g_k^{a=0}\cdot w_k)^2$
    \STATE $\hat{c}_k^{a=1} = (g_k^{a=1}\cdot w_k)^2$
    \ENDFOR
    \STATE $\vec{\mathbf{c}}_l^{a=0} = [\hat{c}_0^{a=0}, \hat{c}_1^{a=0}, ..., \hat{c}_{|\mathcal{K}_l|}^{a=0}]$
    \STATE $\vec{\mathbf{c}}_l^{a=1} = [\hat{c}_0^{a=1}, \hat{c}_1^{a=1}, ..., \hat{c}_{|\mathcal{K}_l|}^{a=1}]$
    \ENDFOR
    \STATE $\mathcal{L}_{d_F} = - \sum_{l=0}^L \mathrm{cos}(\vec{\mathbf{c}}_l^{a=0}, \vec{\mathbf{c}}_l^{a=1})$;  
    % \STATE $\mathcal{L} = \mathcal{L}_{\text{cls}}(\text{F}(\mathbf{X}_0),\mathbf{Y}_0) + \mathcal{L}_{\text{cls}}(\text{F}(\mathbf{X}_1),\mathbf{Y}_1) + \lambda \mathcal{L}_{fair} - \beta \sum_{l=0}^L \mathrm{cos}(\vec{\mathbf{c}}_l^{a=0}, \vec{\mathbf{c}}_l^{a=1})$; \\
    \STATE $\mathcal{L} = \mathcal{L}_{\text{cls}} + \lambda \mathcal{L}_{fair} + \beta \mathcal{L}_{d_F}$; \\
    \STATE $w = w - \eta \nabla_{w}\mathcal{L},\forall w\in \mathcal{W}.$
    \ENDFOR
    \end{algorithmic}
\end{algorithm}

We detail the whole training process in Algorithm \ref{alg:alg}. 
In particular, given a training dataset $\mathcal{D}$, we first sample two groups of samples (ie, $(\mathbf{X}_0, \mathbf{Y}_0)$ and $(\mathbf{X}_1, \mathbf{Y}_1)$) from the two subgroups in the dataset, respectively (See lines 4 and 5). Then, we calculate the cross-entropy loss for both sample groups (See line 7) and calculate the fairness regularization loss (\ie, $\mathcal{L}_\text{fair}=\Delta \text{DP}(\mathbf{F},\mathbf{X}_0,\mathbf{X}_1)$).
% that could be also replaced with the $\Delta \text{EO}$ in \reqref{eq:EO}.
%
After that, we can calculate the gradient of each parameter (\ie, neuron $w_k$) \wrt the classification loss (See lines 11 and 12) and calculate the decision rationale for each neuron and layer (See lines 16 and 17). 
Finally, we calculate the cosine similarity between $\vec{\mathbf{c}}_l^{a=0}$ and $\vec{\mathbf{c}}_l^{a=1}$ and use the whole loss to update the parameters. We defer the algorithm depiction for the EO metric to the Appendix \response{(\ref{app:algorithm_eo})}.

\section{Experiments}

% In this section, we perform experiments to demonstrate the effectiveness of our gradient-guided parity alignment method. Moreover, we perform a detailed analysis to demonstrate the connection between the decision rationale misalignment and model over-parameterization. 

% Meanwhile, the experimental results show that the constraint \textbf{NIA} still suffers from overfitting. We do structural pruning to present that pruning could curb the overfitting of the empirical errors and fairness constraints. 
\vspace{-6pt}
\subsection{Experimental Setup}
%\subsubsection{Datasets, Models and Metrics}

\textbf{Datasets.} In our experiments, we use two tabular benchmarks (\textbf{Adult} and \textbf{Credit}) and one image dataset (\textbf{CelebA}) that are all for binary classification tasks:
    \ding{182} {Adult}~\citep{adult}. The original aim of the dataset Adult is to determine whether a person makes salaries over 50K a year. We consider \textit{gender} as the sensitive attribute, and the vanilla training will lead the model to predict females to earn less salaries.
    \ding{183} {CelebA}~\citep{liu2015faceattributes}. The CelebFaces Attributes dataset is to predict the attributes of face. We split into two subgroups according to the attribute \textit{gender}. Here we consider two attributes classification tasks. For the task to predict whether the hair in an image is \textit{wavy} or not, the standard training will show discrimination towards the male group; when predicting whether the face is \textit{attractive}, the standard training will result in a model prone to predict males as less attractive.
    \ding{184} {Credit}~\citep{credit}.  This dataset is to give an assessment of credit based on personal and financial records. In our paper, we take the attribute \textit{gender} as the sensitive attribute.

\textbf{Models.} For tabular benchmarks, we use the MLP (multilayer perception) \citep{bishop1996neural} as the classification model, which is commonly adopted in classifying tabular data. For the CelebA dataset, we use AlexNet \citep{Alex2012ImageNet} and ResNet-18 \citep{resnet}, both of which are popular in classifying image data \citep{alom2018history}. We mainly show the experimental results of predicting \textit{wavy hair} using AlexNet. More results are in the Appendix (\ref{app:moreresults}).
%The details about the models including the training settings and detailed architectures of the models are given in Appendix.
% For Biased MNIST, we also use MLP as the classification model.

\textbf{Metrics.}
For fairness evaluation, we take two group fairness metrics $\text{DP}$ and $\text{EO}$ as we introduced in the Sec. \ref{sec:fairess} and define $-\text{DP}$ and $-\text{EO}$ as fairness scores since smaller $\text{DP}$ and $\text{EO}$ mean better fairness.
We use the average precision (\text{AP}) for classification accuracy evaluation.
A desired fairness method should achieve smaller $\text{DP}$ or $\text{EO}$ but higher AP (\ie, the top left corner in \figref{fig:fairnes_result}).
\response{Consistent with the previous work \cite{fairmixup,Du2021FairnessVR}, we consider the DP and EO metrics in our work. Moreover, we also explore the Equality of Opportunity \cite{hardt2016equality}
and Predictive Parity \cite{chouldechova2017fair}
. The details are  deferred to the Appendix (\ref{app:eop_pp}).}

%\subsubsection{Baselines}
\textbf{Baselines.} Following the common setups in \citet{fairmixup}, we compare our method with several baselines which are shown to be among the most effective and typical methods:
    \ding{182} Standard training based on empirical risk minimization (ERM) principle (\ie, $\text{w.o.FairReg}$). DNNs are trained only with the cross entropy loss.
    \ding{183} {Oversample (\ie, $\text{w.o.FairReg-Oversample}$)} \citep{stra}. This method samples from the subgroup with rare examples more often, making a balanced sampling in each epoch. 
    % However, the amount of information provided by the different subgroups is still not balanced. 
    % This method easily suffers from overfitting~\citep{stra}, and we denote it as "w.o.FairReg - Oversample".
    % This method is identical to ERM when ERM is satisfied for each subgroup. 
    \ding{184} $\text{FairReg}(\Delta \text{DP}~\text{or}~\Delta \text{EO}, \text{noAug})$ \citep{Madras2018LearningAF}. This method is to directly regularize the fairness metrics, \ie, $\Delta \text{DP}$ or $\Delta \text{EO}$. 
    % Only training by ERM to minimize the loss could also satisfy the fairness metrics constraints when ERM is the optimal solution for each subgroup. Therefore, this method also easily suffers from overfitting.
    %
    \ding{185} $\text{FairReg}(\Delta \text{DP}~\text{or}~\Delta \text{EO}, \text{Aug})$ (\ie, FairMixup) \citep{fairmixup}. This method regularizes the models on paths of interpolated samples between subgroups to achieve fairness. 
    \response{\ding{186} Adversarial \citep{adv}. This method minimizes the adversary's ability to predict sensitive attributes. }
\subsection{Fairness Improvement Performance}
%--------------------------------------------------------
\begin{figure*}
    \centering
    \includegraphics[width=1.0\linewidth]{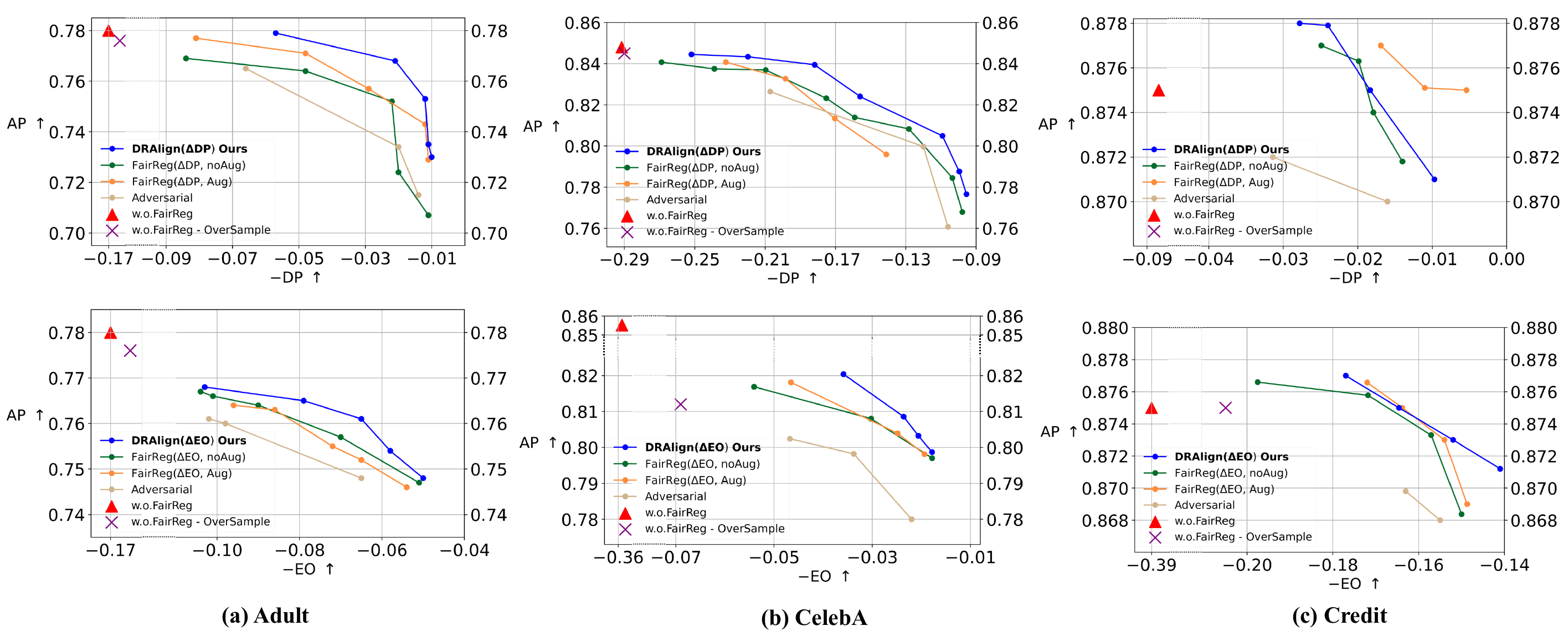}
    \caption{ Comparing different methods on AP vs. (-DP/-EO). According to the common setups, we evaluate $\Delta \text{DP}$-based and $\Delta \text{EO}$-based methods via -DP and -EO, respectively. \response{The plot is drawn by adjusting the hyperparameter $\lambda$ and $\beta$ in Eq. \ref{eq:LossDRA} and Eq. \ref{eq:LossDRA_update}. The detailed hyperparameter settings are in the Appendix (\ref{app:trainingdetails})}. We train networks with the compared methods for 10 times and the averaging results are reported. \response{We show our results are statistically significant via t-test (\ref{app:ttest}).}
    % Our method generally outperforms all the baseline methods in terms of trade-off. For CelebA dataset, we here show the result when predicting \textit{wavy hair}. Note that there is a certain level of variance for each baseline method, we here report the average result over 10 times.
    }
    % \vspace{-0.1in}
    \label{fig:fairnes_result} 
\end{figure*}
%--------------------------------------------------------
\response{As shown in \figref{fig:fairnes_result}, we have following observations:
\ding{182} With the Adult and CelebA datasets, our method (\ie, DRAlign) achieves higher fairness (\ie, higher -DP or -EO scores) than all baseline methods when they have similar AP scores.}
In particular, on the Adult dataset, DRAlign has relative $41.6\%$ DP improvement over the second best method (\ie, $\text{FairReg}(\Delta \text{DP}, \text{Aug})$) when both get around 0.770 AP. 
Overall, our method can enhance the fairness significantly with much less precision sacrifice.
\ding{183} \response{Data augmentation method does not always improve DNN's fairness. For example, on the dataset CelebA, $\text{FairReg}(\Delta \text{DP}, \text{noAug})$ presents slightly higher fairness score (\ie, higher -DP) than $\text{FairReg}(\Delta \text{DP}, \text{Aug})$.}
A potential reason is that the augmented data becomes less realistic due to the rich information in the image modality, which leads to less effective learning.
\ding{184} Although oversampling could improve fairness to some extent, it is less effective than the fairness regularization-based methods (\ie, $\text{FairReg}(*, \text{noAug})$). 
\response{For example, on the CelebA dataset, $\text{w.o.FairReg-Oversample}$ only obtains -0.069 -EO score with the 0.812 AP score, while $\text{FairReg}(\Delta \text{EO}, \text{noAug})$ achieves the -0.054 -EO score with 0.817 AP score. The networks trained by $\text{FairReg}(\Delta \text{EO}, \text{noAug})$ are not only fairer but also of higher accuracy. On the tabular dataset, $\text{w.o.FairReg-Oversample}$ outperforms the $\text{w.o.FairReg}$ by a small margin. }
\ding{185} On the Credit dataset, FairReg($\Delta \text{DP}$, Aug) achieves better results than DRAlign under the DP metric although our method still outperforms the regularization-based one.
A potential reason is that the data size of the Credit is small (\ie, 500 training samples) and the data augmentation can present obvious advantages by enriching the training data significantly. \response{The data augmentation and our decision rationale alignment are two independent ways to enhance fairness. Intuitively, we can combine the two solutions straightforwardly. We did further experiments and found that our DRAlign could still improve FairReg($\Delta \text{DP}$, Aug). In addition, our experiments show that the decision rationale alignment itself could still slightly improve fairness when the fairness regularization item ($\mathcal{L}_\text{fair}(\text{F})$) is removed. More details are put in the Appendix (\ref{app:dataaug},\ref{app:nofairreg}).}
%
% The cause might be that oversampling only increases the number of training examples only while the information provided by training samples of different subgroups is still not equal.

\subsection{Discussion and Analysis}

% --------------------------------------------------
\begin{figure*}[htb]
% \vspace{-30pt}  
\centering
\includegraphics[width=0.8\linewidth]{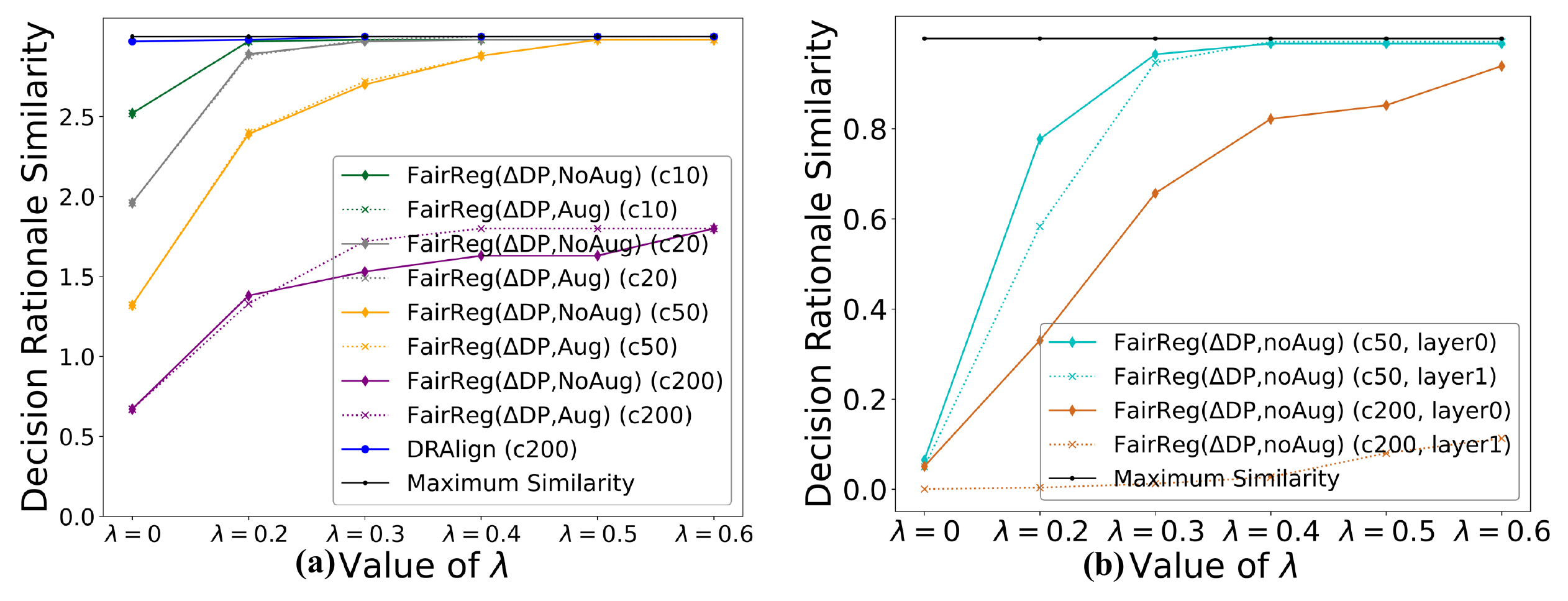}
% \vspace{10pt}   
\caption{ (a) : correlation between $\lambda$ and decision rationale similarity score. (b) layer-wise analysis for correlation between $\lambda$ and decision rationale similarity score. 
% \text{FairReg($\Delta DP$,NoAug) (c50, layer0)} denotes the decision rationale similarity of the first layer in the model with the architecture of 50 hidden channels.
}
\label{fig:op_ana}
\label{fig:op_ana}
% \end{figure}
\end{figure*}

\textbf{Connection with over-parameterization.} 

To better understand the cause of the decision rationale misalignment, we further investigate the connection between decision rationale misalignment and model over-parameterization. 
We conduct an empirical study on the Adult dataset using 3-layer MLP networks based on $\text{FairReg}(\Delta \text{DP}, \text{noAug})$. Specifically, we explore 4 MLP architectures, where the hidden sizes are set as 10, 20, 50, and 200, respectively. The corresponding parameter sizes of the 4 networks are 1331, 2861, 8651, and 64601.
For each architecture, we draw a plot \wrt different $\lambda$ for $\text{FairReg}(\Delta \text{DP}, \text{noAug})$ to show the decision rationale similarity score (\ie, $\sum_{l=0}^L \text{cos}(\vec{\mathbf{c}}_l^{a=0}, \vec{\mathbf{c}}_l^{a=1})$ in \secref{subsec:gradient-guided}). 
We denote the four trained models as FairReg($\Delta \text{DP}$,noAug) (c10), FairReg(noAug) (c20), FairReg($\Delta \text{DP}$,noAug) (c50), and FairReg($\Delta \text{DP}$,noAug) (c200), respectively, according to their hidden sizes. 
\response{\emph{Accuracy performance of these models and more results under $\Delta EO$ metric are put in the Appendix (\ref{app:aps},\ref{app:overparameter_eo}).}}
With \figref{fig:op_ana}~(a), we have the following observations:
\ding{182} The decision rationale similarity consistently ascends when $\lambda$ increases. When $\lambda$ becomes 0.5, decision rationale similarities of FairReg($\Delta \text{DP}$,noAug) (c10), FairReg($\Delta \text{DP}$,noAug) (c20)and FairReg($\Delta \text{DP}$,noAug) (c50) reach the same maximum score (\ie, 3.0 for any 3-layer MLP network). We conclude that larger $\lambda$ (stricter fairness constraint) results in a higher decision rationale similarity.
\ding{183} The misalignment of decision rationale is more likely to occur in the over-parameterized networks. For the largest network FairReg($\Delta \text{DP}$,noAug) (c200), even when the $\lambda$ is set as 0.6 for a strict fairness constraint, the decision rationale similarity score only reaches 1.8 which is much smaller than the values on other variants and infers that the decision rationale is still far from being aligned.

Furthermore, we also report the results of augmentation-based method, \ie, FairReg($\Delta \text{DP}$,Aug). We find that data augmentation can generally mitigate the misalignment but still fails to completely align the decision rationale (See the plot of FairReg($\Delta \text{DP}$,noAug) (c200)).
Our method DRAlign is able to achieve the maximum similarity on all $\lambda$ settings even on the architecture with hidden size 200. (See the plot of DRAlign (c200))
This enlightens us that common methods such as data augmentation that aims to address over-parameterization can not completely solve the misalignment, while our gradient-guided parity alignment method can directly improve the alignment.

% %-----------------------------------------------------
% \begin{wrapfigure}{R}{0.45\textwidth}
% \centering
% \includegraphics[width=1\linewidth]{pics/ana.png}
% \vspace{-10pt}
% \caption{Analysis for correlation between $\lambda$ and Decision similarity score. $\text{FairReg(NoAug)}_{c10}$ is the model trained by FairReg method using the architecture of 10 hidden channels.}
% \vspace{-0.1in}
% \label{fig:ana}
% \end{wrapfigure}
% %-----------------------------------------------------

\textbf{Layer-wise decision rationale alignment analysis.}

We further conduct a layer-wise analysis to understand which layer owns better decision rationale alignment. We calculate the decision rationale similarity for the $1$st and $2$nd layer (\ie, $\text{cos}(\vec{\mathbf{c}}_{l=0}^{a=0}, \vec{\mathbf{c}}_{l=0}^{a=1})$ and $\text{cos}(\vec{\mathbf{c}}_{l=1}^{a=0}, \vec{\mathbf{c}}_{l=1}^{a=1})$). From \figref{fig:op_ana}~(b), we see that: for both layers, the layer-wise similarity score ascends when $\lambda$ increases. This is consistent with the observation that stricter fairness constraint results in a higher decision rationale similarity. 
As we compare the $1$st and $2$nd layers, we can observe that the similarity score of the first layer is generally higher. Moreover, we can see that for smaller models (\ie, models with hidden size 50), the similarity gap between the first layer and the second layer is relatively trivial. However, for models with hidden size 200, the similarity score of the second layer is rather low (\ie, the score is 0.113 even when the $\lambda$ is 0.6). Thus, the misalignment of the deeper layer is severer.

\section{\response{Related Work}}

% 1, individual and group
% 2, mitigation
% 3, 

% \subsection{Fairness in Deep Learning}
{\bf Fairness in Deep Learning.}
Deep learning models, while potent, often display questionable behavior in relation to critical issues such as robustness, privacy, and trustworthiness \cite{goodfellow2014explaining,kurakin2018adversarial,Liu2019Perceptual,Liu2020Biasbased,Liu2020Spatiotemporal,liu2021ANP,Liu2023Xadv,guo2023towards,huang2023robustness}. Among these concerns, discrimination stands out as a highly influential factor with considerable social implications. \response{There are different methods to evaluate fairness in deep learning, among which individual fairness~\citep{ADF,EIDIG,vif,xiao2023latent}, group fairness~\citep{gf1,gf2,gf3,Garg2020Fairness}, and counterfactual fairness~\citep{Kusner2017Counterfactual} are the mainstream. }
% Individual fairness follows the philosophy that similar inputs should yield similar predictions. The definition of input similarity is an open question in individual fairness issues. Group fairness is derived by calculating and comparing the predictions for each group. Counterfactual fairness considers whether a decision is fair toward an individual in a causal sense, which is also an individual-level definition. As group fairness is more widely adopted in fairness research \cite{fair_review}, we focus on group fairness which is derived by calculating and comparing the predictions for each group. }
We focus on group fairness which is derived by calculating and comparing the predictions for each group
\response{There is a line of work dedicated to alleviating unjustified bias. For example, \citet{stra} compare mitigation methods including oversampling, adversarial training, and other domain-independent methods. Some work proposes to disentangle unbiased representations to ensure fair DNNs. On the contrary, \citet{Du2021FairnessVR} directly repair the classifier head even though the middle representations are still biased. To improve fairness, it is also popular to constrain the training process by imposing regularization. \citet{pmlr-v65-woodworth17a} regularize the covariance between predictions and sensitive attributes. \citet{Madras2018LearningAF} relax the fairness metrics for optimization. Although such methods are easy to be implemented and integrated into the training process, these constraints suffer from overfitting~\citep{overfitting}. The model with a large number of parameters could memorize 
% the data patterns of 
the training data, which causes the fairness constraints to fit well only in the training process.
% while violated on the test data.
\citet{fairmixup} ensure better generalization via data augmentation (\eg, mix-up) to reduce the trade-off between fairness and accuracy.
}
% There are different methods to evaluate fairness in deep learning, among which individual fairness~\citep{ADF,EIDIG,vif} and group fairness~\citep{gf1,gf2,gf3} are the mainstream. We focus on group fairness which is derived by calculating and comparing the predictions for each group. There is a line of work dedicated to alleviating unjustified bias. For example, \citet{stra} compare mitigation methods including oversampling, adversarial training, and other domain-independent methods. Some work proposes to disentangle unbiased representations to ensure fair DNNs. On the contrary, \citet{Du2021FairnessVR} directly repair the classifier head even though the middle representations are still biased. To improve fairness, it is also popular to constrain the training process by imposing regularization. \citet{pmlr-v65-woodworth17a} regularize the covariance between predictions and sensitive attributes. \citet{Madras2018LearningAF} relax the fairness metrics for optimization. Although such methods are easy to be implemented and integrated into the training process, these constraints suffer from overfitting~\citep{overfitting}. The model with a large number of parameters could memorize the data patterns of the training data, which causes the fairness constraints to fit well in the training process while violated on the test data. \citet{fairmixup} ensure better generalization via data augmentation (\eg, mix-up) to reduce the trade-off between fairness and accuracy. 
However, these methods barely pay attention to the rationale behind the fair decision results. Besides, some studies propose measures for procedural fairness that consider the input features used in the decision process and evaluate the moral judgments of humans regarding the use of these features \cite{grgic2016case,grgic2018beyond}. They focus on feature selection for procedurally fair learning. In this paper, we further analyze the decision rationales behind the fair decision results in the training process and reveal that ensuring the fair decision rationale could further improve fairness. 

% \subsection{Understanding DNNs Decision Rationale via Decomposing}
{\bf Understanding DNNs Decision Rationale.}
There are some interpretable methods enabling DNNs models to present their behaviors in understandable ways to humans~\citep{inter1,inter2,inter3,liang2019knowledge,zhang2020interpreting,li2021understanding}. Specifically, there is a line of work that depicts the decision rationale of DNNs via neuron behaviors analysis. Routing paths composed of the critical nodes (\eg neurons with the most contribution to the final classification on each layer) can be extracted in a learnable way to reflect the network's semantic information flow regarding to a group of data~\citep{khakzar2021neural}. Conquering the instability existing in the learnable method, \citet{advprofile} propose an activation-based back-propagation algorithm to decompose the entire DNN model into multiple components composed of structural neurons. \response{Meanwhile, ~\citet{npc} base the model function analysis on the neuron contribution calculation and reveal that the neuron contribution patterns of OOD samples and adversarial samples are different from that of normal samples, resulting in wrong classification results. \citet{Zheng2022NeuronFair} analyze neurons sensitive to individual discrimination and generate testing cases according to sensitive neuron behaviors.
% Moreover, \cite{Tian2022NeuronCD} improve the generalization capability by maximizing the amount of activated neurons of the model. 
However, these methods analyze neuron behaviors via static analysis or in a learnable way. These analysis methods result in huge time overhead, making their integration into the training process difficult, which restricts their applications in optimizing the training.}

In our paper, we follow the spirit of analyzing neuron behaviors to understand the model decision rationale.
% for a group of data. 
Unlike previous methods, our method successfully simplifies the estimation process of neuron contribution and can be easily integrated into the training process to optimize the model.

\section{Conclusions and Future Work}

In this work, we have studied the fairness issue of deep models from the perspective of decision rationale and defined the \textit{parameter parity score} to characterize the decision rationale shifting across subgroups. We observed that such a decision rationale-aware characterization has a high correlation to the fairness of deep models, which means that a fairer network should have aligned decision rationales across subgroups. To this end, we formulated fairness as the decision rationale alignment (DRAlign) and proposed the \textit{gradient-guided parity alignment} to implement the new task. The results on three public datasets demonstrate the effectiveness and advantages of our methods and show that DRAlign is able to achieve much higher fairness with less precision sacrifice than all existing methods. 

Although promising, our method also presents some drawbacks: (1) it requires the computation of second-order derivatives; and (2) the gradient-guided parity alignment method is limited to the layer-wise DNN architecture. In the future, we are interested in solving these limitations.

\section*{Acknowledgments}
This research/project is supported by the National Research Foundation, Singapore
under its AI Singapore Programme (AISG Award No: AISG2-PhD-2021-08-022T).
It is also supported by A*STAR Centre for Frontier AI Research, the National Research Foundation Singapore and DSO National Laboratories under the AI Singapore Programme (AISG Award No: AISG2-RP-2020-019), National Satellite of Excellence in Trustworthy Software System No.~NRF2018NCR-NSOE003-0001, NRF Investigatorship No.~NRF-NRFI06-2020-0001, and the National Natural Science Foundation of China 62206009. We gratefully acknowledge the support of NVIDIA AI Tech Center (NVAITC). %It is also supported by SKLOIS, Institute of Information Engineering, Chinese Academy of Sciences, China. 

\bibliography{ref}
\bibliographystyle{icml2023}

% \appendix
% \section{Appendix}

\newpage
\appendix
\onecolumn

\section{Appendix}

\subsection{Training Details}
\label{app:trainingdetails}
\textbf{Adult Dataset.} 
The parameter setting of the Adult Dataset is shown in Table \ref{tab:adult}. We follow the settings in \citet{fairmixup} for data preprocessing. The hidden size of MLP is 200. \response{We use Adam as the learning optimizer and the batch size is set as 1000 for the DP metric and 2000 for the EO metric following the setting in \citet{fairmixup}.}

\begin{table}[h]
    \centering
    \caption{Setting for Adult Dataset training with MLP.} 
    % \begin{adjustbox}{max width=\textwidth,keepaspectratio}
    \resizebox{1.0\columnwidth}{!}{%
    \begin{tabular}{c|c|c|c|c|c}
    \toprule
        % Decision rationale shifting ($d_{\text{F}}$) & $0.6236$  & $0.391 \pm 0.338$ & $0.101 \pm 0.084$ & $0,070 \pm 0.045$  & $0.0458 \pm 0.023$   & $0.039 \pm 0.029$ \\ \midrule
        & w.o.FairReg & w.o.FairReg - OverSample& FairReg(*, noAug) & FairReg(*, Aug) & DRAlign \\ \midrule
        Training Epochs for DP & 20 & 20 & 20 & 20  & 20  \\ \midrule
        Training Epochs for EO & 20 & 20 & 20 & 20  & 20  \\ \midrule
        Learning rate  & 0.001 & 0.001 & 0.001 & 0.001  & 0.001  \\ \midrule
        Range of $\lambda$ for DP & - & - & [0.2,0.3,0.4,0.5,0.6]& [0.2,0.3,0.4,0.5,0.6] & [0.1,0.2,0.3,0.4,0.5]  \\  \midrule
        $\beta$ for DP& -  & - & - &  - & $\lambda/10$ \\ \midrule
        Range of $\lambda$ for EO & - & - & [0.5\textasciitilde2.0]& [0.5\textasciitilde2.0] & [0.5\textasciitilde2.0] \\  \midrule
        $\beta$ for EO & -  & - & - &  - & $\lambda/10$ \\
         \bottomrule 
    \end{tabular}
    }
    % \end{adjustbox}
% \vspace{-10pt}
\label{tab:adult}
\end{table}

\textbf{CelebA Dataset.} The parameter setting of the CelebA Dataset is shown in Table \ref{tab:celeba}. We follow the settings in \cite{fairmixup} for data preprocessing. \response{We use Adam as the learning optimizer and the batch size is set as 64 for the DP metric and 128 for the EO metric following the setting in \citet{fairmixup}.}
\begin{table}[h]
    \centering
    \caption{Setting for CelebA Dataset training with AlexNet.} 
    % \begin{adjustbox}{max width=\textwidth,keepaspectratio}
    \resizebox{1.0\columnwidth}{!}{%
    \begin{tabular}{c|c|c|c|c|c}
    \toprule
        % Decision rationale shifting ($d_{\text{F}}$) & $0.6236$  & $0.391 \pm 0.338$ & $0.101 \pm 0.084$ & $0,070 \pm 0.045$  & $0.0458 \pm 0.023$   & $0.039 \pm 0.029$ \\ \midrule
        & w.o.FairReg & w.o.FairReg - OverSample& FairReg(*, noAug) & FairReg(*, Aug) & DRAlign \\ \midrule
        Training Epochs for DP & 15 & 15 & 15 & 30  & 15  \\ \midrule
        Training Epochs for EO & 30 & 30 & 30 & 60  & 30  \\ \midrule
        Learning rate  & 0.0001 & 0.0001 & 0.0001 & 0.0001  & 0.0001  \\ \midrule
        Range of $\lambda$ for DP & - & - & [0.1\textasciitilde0.7]& [0.1\textasciitilde0.6] & [0.1\textasciitilde0.7]  \\  \midrule
        $\beta$ for DP& -  & - & - &  - & 0.01 \\ \midrule
        Range of $\lambda$ for EO & - & - & [0.1,0.4,0.7]&  [0.1,0.4,0.7] &  [0.1,0.4,0.7,1.0]\\  \midrule
        $\beta$ for EO & -  & - & - &  - & 0.01 \\
         \bottomrule 
    \end{tabular}
    }
    % \end{adjustbox}
% \vspace{-10pt}
\label{tab:celeba}
\end{table}

\textbf{Credit Dataset.} The parameter setting of the Credit Dataset is shown in Table \ref{tab:german}. We follow the settings in \cite{Peixin2020White} for data preprocessing. \response{We use Adam as the learning optimizer and the batch size is set as 400 for the DP metric and 500 for the EO metric.}
\begin{table}[h]
    \centering
    \caption{Setting for Credit Dataset.} 
    % \begin{adjustbox}{max width=\textwidth,keepaspectratio}
    \resizebox{1.0\columnwidth}{!}{%
    \begin{tabular}{c|c|c|c|c|c}
    \toprule
        % Decision rationale shifting ($d_{\text{F}}$) & $0.6236$  & $0.391 \pm 0.338$ & $0.101 \pm 0.084$ & $0,070 \pm 0.045$  & $0.0458 \pm 0.023$   & $0.039 \pm 0.029$ \\ \midrule
        & w.o.FairReg & w.o.FairReg - OverSample& FairReg(*, noAug) & FairReg(*, Aug) & DRAlign \\ \midrule
        Training Epochs for DP & 20 & 20 & 20 & 20  & 20  \\ \midrule
        Training Epochs for EO & 20 & 20 & 20 & 20  & 20  \\ \midrule
        Learning rate  & 0.001 & 0.001 & 0.001 & 0.001  & 0.001  \\ \midrule
        Range of $\lambda$ for DP & - & - & [0.2,0.8,1.0,2.0]& [0.2,0.8,2.0] & [0.8,1.0,2.0,3.0]  \\  \midrule
        $\beta$ for DP& -  & - & - &  - & 0.005 \\ \midrule
        Range of $\lambda$ for EO & - & - & [0.2,0.4,0.6,0.8]& [0.2,0.4,0.8,1.0] & [0.6, 0.8,1.0,2.0] \\  \midrule
        $\beta$ for EO & -  & - & - &  - & 0.01 \\
         \bottomrule 
    \end{tabular}
    }
    % \end{adjustbox}
% \vspace{-10pt}
\label{tab:german}
\end{table}

\response{In our paper, we did a rough search for the hyper-parameter $\beta$. Taking CelebA dataset as an example, we mainly search $\beta$ value in the range {0.001, 0.01, 0.1}. When $\beta$ is set as 0.001, the training process is close to that of FairReg, which means that our decision rationale alignment item is ignored in the training because $\beta$ is too small. When $\beta$ is 0.1, the training process will optimize the decision rationale alignment first and cause a detrimental influence on the optimization of other loss items. We finally choose 0.01 as the $\beta$ value. We have found that the training process is stable with a proper $\beta$ setting. We train the models for the CelebA dataset from scratch.}

In our paper, we mainly consider the parameters of multiplicative weights (neurons/filters) in the convolution layers and the linear layers, because we focus on the decision rationale which could be defined by the influence of neuron/filter/parameter that is usually regarded as a semantic unit. We did not consider the biases parameters and the parameters in BN layers.

\subsection{Algorithm of DRAlign When Training under The EO Metric}
\label{app:algorithm_eo}
The training algorithm for EO metric is shown in Algorithm \ref{alg:alg_eo}.

\begin{algorithm}[t]
\caption{Gradient-guided Parity Alignment for The EO Metric}
\label{alg:alg_eo}

    \begin{algorithmic}
    
    \STATE {\bfseries Input:} Network $\text{F}$ with parameters $\mathcal{W}=\{w_0,\ldots,w_K\}$, epoch index set $\mathcal{E}$, training data $\mathcal{D}$, 
    % % iteration index set $\mathcal{I}$ per epoch,
    batch size $B$, network layers $L$, neurons in the $l$th layer $\mathcal{K}_l$, hyper-parameters $\lambda$ and $\beta$, learning rate $\eta$
    % \REPEAT
    % \STATE Initialize $noChange = true$.
    \FOR{$e \in \mathcal{E}$}
    \STATE // Sampling $B$ samples from subgroups in $\mathcal{D}$
    \STATE $[\mathbf{X_{00}}, \mathbf{Y_{00}}]$  = Sample($D$, a=0, y=0, $B$);
   \\
    \STATE $[\mathbf{X_{01}}, \mathbf{Y_{01}}]$  = Sample($D$, a=0, y=1, $B$);
    \\
    \STATE $[\mathbf{X_{10}}, \mathbf{Y_{10}}]$  = Sample($D$, a=1, y=0, $B$);
    \\
    \STATE $[\mathbf{X_{11}}, \mathbf{Y_{11}}]$  = Sample($D$, a=1, y=1, $B$);
    \STATE // Calculating loss and updating the model
    \STATE $\mathcal{L}_c = \mathcal{L}_{\text{cls}}(\text{F}(\mathbf{X_{00}}),\mathbf{Y_{00}})) + \mathcal{L}_{\text{cls}}(\text{F}(\mathbf{X_{01}}),\mathbf{Y_{01}}) + \mathcal{L}_{\text{cls}}(\text{F}(\mathbf{X_{10}}),\mathbf{Y_{10}})) + \mathcal{L}_{\text{cls}}(\text{F}(\mathbf{X_{11}}),\mathbf{Y_{11}})$; 
    \STATE $\mathcal{L}_\text{fair} =\Delta \text{EO}(\mathbf{F},\mathbf{X}_{00},\mathbf{X}_{01},\mathbf{X}_{10},\mathbf{X}_{11})$
    \FOR{$l \in \mathcal{L}$}
    \FOR{$k \in \mathcal{K}_l$}
    \STATE $g_k^{a=0,y=0} =  \frac{\partial(\mathcal{L}_{\text{cls}}(\text{F}(\mathbf{X_{00}}),\mathbf{Y_{00}}))}{\partial w_k}$;
    % \\ 
    \STATE  $g_k^{a=1,y=0} = \frac{\partial(\mathcal{L}_{\text{cls}}(\text{F}(\mathbf{X_{10}}),\mathbf{Y_{10}}))}{\partial w_k}$; 
    \\
    \STATE  $g_k^{a=0,y=1} =  \frac{\partial(\mathcal{L}_{\text{cls}}(\text{F}(\mathbf{X_{01}}),\mathbf{Y_{01}}))}{\partial w_k}$;
    % \\ 
    \STATE  $g_k^{a=1,y=1} = \frac{\partial(\mathcal{L}_{\text{cls}}(\text{F}(\mathbf{X_{11}}),\mathbf{Y_{11}}))}{\partial w_k}$; 
    \\
    \STATE  $\hat{c}_k^{a=0,y=0} = (g_k^{a=0,y=0}\cdot w_k)^2$;
    % \\
    \STATE  $\hat{c}_k^{a=1,y=0} = (g_k^{a=1,y=0}\cdot w_k)^2$; 
    \\
    \STATE  $\hat{c}_k^{a=0,y=1} = (g_k^{a=0,y=1}\cdot w_k)^2$;
    % \\
    \STATE  $\hat{c}_k^{a=1,y=1} = (g_k^{a=1,y=1}\cdot w_k)^2$; 
    \ENDFOR
    \STATE $\vec{\mathbf{c}}_l^{a=0,y=0} = [\hat{c}_0^{a=0,y=0}, \hat{c}_1^{a=0,y=0}, ..., \hat{c}_{\mathcal{K}_l}^{a=0,y=0}]$;
    \\
    \STATE $\vec{\mathbf{c}}_l^{a=1,y=0} = [\hat{c}_0^{a=1,y=0}, \hat{c}_1^{a=1,y=0}, ..., \hat{c}_{\mathcal{K}_l}^{a=1,y=0}]$;
    \\
    \STATE $\vec{\mathbf{c}}_l^{a=0,y=1} = [\hat{c}_0^{a=0,y=1}, \hat{c}_1^{a=0,y=1}, ..., \hat{c}_{\mathcal{K}_l}^{a=0,y=1}]$;
    \\
    \STATE $\vec{\mathbf{c}}_l^{a=1,y=1} = [\hat{c}_0^{a=1,y=1}, \hat{c}_1^{a=1,y=1}, ..., \hat{c}_{\mathcal{K}_l}^{a=1,y=1}]$;
    \STATE $\mathcal{L}_{d_F} = - \sum_{l=0}^L \mathrm{cos}(\vec{\mathbf{c}}_l^{a=0,y=0}, \vec{\mathbf{c}}_l^{a=1,y=0}) + \sum_{l=0}^L \mathrm{cos}(\vec{\mathbf{c}}_l^{a=0,y=1}, \vec{\mathbf{c}}_l^{a=1,y=1})$; 
    \STATE $\mathcal{L} = \mathcal{L}_c + \lambda \mathcal{L}_{fair} + \beta \mathcal{L}_{d_F}$;  \\
    \STATE $w \leftarrow w - \eta \nabla_{w}\mathcal{L},\forall w\in \mathcal{W}.$
    \ENDFOR
    \ENDFOR
    \end{algorithmic}
\end{algorithm}

\subsection{More Experimental Results}
\label{app:moreresults}
\subsubsection{Classification for Attractive Attribute}
In our paper, on the CelebA dataset, we show the results of predicting \textit{wavy hair} attribute. Here, we also show the results of classifying \textit{attractive} attribute adopting AlexNet. For better observation, we show our results in Table \ref{tab:r1}. We find that our method outperforms FairReg(noAug) both in AP and in the fairness metric.

\begin{table}[h]
    \centering
    \caption{Comparison between DRAlign(ours) and FairReg(*, noAug) when classifying \textit{attractive} attribute.} 
    % \begin{adjustbox}{max width=\textwidth,keepaspectratio}
    \resizebox{0.8\columnwidth}{!}{%
    \begin{tabular}{c|c|c|c|c|c|c|c}
    \toprule
        & \multicolumn{3}{| c |}{$-$DP} & \multicolumn{3}{| c |}{$-$EO} \\ 
        & $\lambda=0.1$ & $\lambda=0.2$ & $\lambda=0.3$ & $\lambda=0.1$ & $\lambda=0.4$ & $\lambda=1.0$ \\ \midrule
        $\text{AP}_{DRAlign}$ & \textbf{0.8956} & \textbf{0.8895} & 0.8783 & \textbf{0.8760}  & \textbf{0.8735} &  \textbf{0.8717}  \\ \midrule
        $\text{Fairness}_{DRAlign}$ & $\textbf{-0.3196}$ & $\textbf{-0.2727}$ & $\textbf{-0.2126}$ & $\textbf{-0.0520}$  & $\textbf{-0.0337}$ &  $\textbf{-0.0243}$  \\ \midrule
        $\text{AP}_{FairReg(*,noAug)}$ & 0.8942 & 0.8873  & \textbf{0.8807} & 0.8733  & 0.8707 & 0.8681   \\ \midrule
        $\text{Fairness}_{FairReg(*,noAug)}$ & $-0.3305$ &$ -0.2819$  & $-0.2377$ & $-0.0533$  & $-0.0374$ & $-0.0273 $  \\ \midrule
         \bottomrule 
    \end{tabular}
    }
    % \end{adjustbox}
% \vspace{-10pt}
\label{tab:r1}
\end{table}

\subsubsection{Classification for Wavy Hair Based on ResNet-18}
In our algorithm, we expect to reduce the parity score for all layers. However, for some larger architectures such as ResNet-18, it is relatively difficult to optimize all layers. To address such a problem, we here only align the last two layers. We find that only aligning the last two layers could also improve fairness. The loss function is revised as follows:
\begin{align} 
\mathcal{L} & = \text{E}_{(\mathbf{x},y) \sim P}(\mathcal{L}_{\text{cls}}(\text{F}(\mathbf{x}),y)) + \lambda \mathcal{L}_\text{fair}(\text{F}) - \beta \sum_{l=L-1}^L \text{cos}(\vec{\mathbf{c}}_l^{a=0}, \vec{\mathbf{c}}_l^{a=1}),
\end{align}
The experimental results are shown in Table \ref{tab:r2}.

\begin{table}[h]
    \centering
    \caption{Comparison between DRAlign(ours) and FairReg(*, noAug) when classifying \textit{Wavy hair} attribute using ResNet-18.} 
    % \begin{adjustbox}{max width=\textwidth,keepaspectratio}
    \resizebox{0.8\columnwidth}{!}{%
    \begin{tabular}{c|c|c|c|c|c|c|c}
    \toprule
        & \multicolumn{3}{| c |}{$-$DP} & \multicolumn{3}{| c |}{$-$EO} \\ 
        & $\lambda=0.1$ & $\lambda=5.0$ & $\lambda=10.0$ & $\lambda=0.2$ & $\lambda=5.0$ & $\lambda=10.0$ \\ \midrule
        $\text{AP}_{DRAlign}$ & \textbf{0.8578} & \textbf{0.8385} & \textbf{0.8179} & \textbf{0.8212} & \textbf{0.7965} & \textbf{0.7703}  \\ \midrule
        $\text{Fairness}_{DRAlign}$ & $\textbf{-0.3011}$ & $\textbf{-0.2723}$ & $\textbf{-0.2481}$ & $\textbf{-0.1294}$  & $\textbf{-0.0495}$ &  $\textbf{-0.0446}$  \\ \midrule
        $\text{AP}_{FairReg(*,noAug)}$ & 0.8506 & 0.8355  & 0.8123 & 0.8063  & 0.7857 & 0.7560   \\ \midrule
        $\text{Fairness}_{FairReg(*,noAug)}$ & $-0.3063$ &$ -0.2795$  & $-0.2552$ & $-0.1832$  & $-0.0498$ & $-0.0494 $  \\ \midrule
         \bottomrule 
    \end{tabular}
    }
    % \end{adjustbox}
\vspace{-10pt}
\label{tab:r2}
\end{table}

\subsection{Connection With Over-parameterization under EO Metric}
\label{app:overparameter_eo}
We here analyze the connection between decision rationale alignment and over-parameterization under EO metric. We show the results on the Adult dataset adopting 3-layer MLP models. The maximum alignment score is 6.0. Here we also conclude that over-parameterization might prevent the alignment of decision rationale and stricter fairness regularizations require fairer decision rationale. 
\begin{table}[h] \

    \centering
    \caption{Connection between decision rationale similarity and over-parameterization under EO metric.} 
    % \begin{adjustbox}{max width=\textwidth,keepaspectratio}
    \resizebox{0.8\columnwidth}{!}{%
    \begin{tabular}{c|c|c|c|c|c}
    \toprule
        & $\lambda=0.5$ & $\lambda=0.8$ & $\lambda=1.0$ & $\lambda=2.0$ & $\lambda=3.0$\\ \midrule
        \text{FairReg($\Delta EO$,noAug)}, (c10) & 6.0 & 6.0 & 6.0 & 6.0  & 6.0  \\ \midrule
        \text{FairReg($\Delta EO$,noAug)}, (c20) & 5.7 & 5.7 & 5.8 & 6.0  & 6.0 \\ \midrule
        \text{FairReg($\Delta EO$,noAug)}, (c50)  & 5.6 & 5.7 & 5.8 & 6.0  & 6.0  \\ \midrule
        \text{FairReg($\Delta EO$,noAug)}, (c200) & 5.6 & 5.7 & 5.8 & 5.9 & 6.0 \\  \midrule
        \text{FairReg($\Delta EO$,Aug)}, (c10)& 6.0 & 6.0 & 6.0 &  6.0 & 6.0 \\ \midrule
        \text{FairReg($\Delta EO$,Aug)}, (c20)& 6.0 & 6.0 & 6.0 & 6.0 & 6.0 \\  \midrule
        \text{FairReg($\Delta EO$,Aug)}, (c50)& 5.9 & 6.0 & 6.0 & 6.0 & 6.0 \\  \midrule
        \text{FairReg($\Delta EO$,Aug)}, (c200)& 5.7 & 5.9 & 6.0& 6.0& 6.0\\  \midrule
         \bottomrule 
    \end{tabular}
    }
    % \end{adjustbox}
\vspace{-10pt}
\label{tab:over}
\end{table}

\subsection{\response{Training Time Estimation}}
\label{app:trainingtime}
\response{We here show the time consumption of different methods on the Adult dataset, CelebA dataset, and Credit dataset in Table ~\ref{tab:time_adult}, Table ~\ref{tab:time_celeba} and Table ~\ref{tab:time_credit} respectively. Please be noted that the FairReg(*,Aug) method also requires the calculation of a second-order derivative. Moreover, as a method based on data augmentation, the FairReg(*,Aug) method requires more time to converge.}

\begin{table}[h] \
    \centering
    \caption{\response{Training time estimation when training with Adult dataset under the DP and EO metric.} }
    % \begin{adjustbox}{max width=\textwidth,keepaspectratio}
    \resizebox{0.8\columnwidth}{!}{%
    \begin{tabular}{c|c|c|c|c|c}
    \toprule
        & w.o.FairReg & w.o.FairReg - OverSample& \text{FairReg(*,noAug)} & \text{FairReg(*,Aug)} & DRAlign \\ \midrule
        DP & 8.2s & 10.1s & 10.5s & 14.7s  & 14.5s  \\ \midrule
        EO & 12.5s & 14.6s & 15.0s & 33.2s  & 30.1s  \\ \midrule
         \bottomrule 
    \end{tabular}
    }
    % \end{adjustbox}
\vspace{-10pt}
\label{tab:time_adult}
\end{table}

\begin{table}[h] \
    \centering
    \caption{\response{Training time estimation when training with CelebA dataset and AlexNet under the DP and EO metric.} }
    % \begin{adjustbox}{max width=\textwidth,keepaspectratio}
    \resizebox{0.8\columnwidth}{!}{%
    \begin{tabular}{c|c|c|c|c|c}
    \toprule
        & w.o.FairReg & w.o.FairReg - OverSample& \text{FairReg(*,noAug)} & \text{FairReg(*,Aug)} & DRAlign \\ \midrule
        DP & 611.3s & 725.2s & 811.6s & 1995.3s  & 1397.8s  \\ \midrule
        EO & 661.8s & 761.8s & 865.8s & 3640.8s  & 3278.2s  \\ \midrule
         \bottomrule 
    \end{tabular}
    }
    % \end{adjustbox}
% \vspace{-10pt}
\label{tab:time_celeba}
\end{table}

\begin{table}[h] \
    \centering
    \caption{\response{Training time estimation when training with Credit dataset under the DP and EO metric.} }
    % \begin{adjustbox}{max width=\textwidth,keepaspectratio}
    \resizebox{0.8\columnwidth}{!}{%
    \begin{tabular}{c|c|c|c|c|c}
    \toprule
        & w.o.FairReg & w.o.FairReg - OverSample& \text{FairReg(*,noAug)} & \text{FairReg(*,Aug)} & DRAlign \\ \midrule
        DP & 6.1s & 8.6s & 8.7s & 12.5s  & 12.1s  \\ \midrule
        EO & 8.5s & 10.7s & 11.1s & 13.5s  & 13.0s  \\ \midrule
         \bottomrule 
    \end{tabular}
    }
    % \end{adjustbox}
\vspace{-10pt}
\label{tab:time_credit}
\end{table}

\subsection{\response{The AP Values of Various Model Architectures}}
\label{app:aps}
\response{Table ~\ref{tab:aps} show the AP values of different model architectures. The model is chosen according to the performance on the validation dataset. We can see that larger models are prone to have higher APs. }
\begin{table}[h] 
    \centering
    \caption{\response{The AP Values of Different Model Architectures.} }
    % \begin{adjustbox}{max width=\textwidth,keepaspectratio}
    \resizebox{0.6\columnwidth}{!}{%
    \begin{tabular}{c|c|c|c|c|c|c|c|c}
    \toprule
        & $\lambda=0$ & $\lambda=0.1$& $\lambda=0.2$ & $\lambda=0.3$ & $\lambda=0.4$ & $\lambda=0.5$ & $\lambda=0.6$ & $\lambda=0.7$\\ \midrule
        $c_{10}$ & 0.781 & 0.780 & 0.776 & 0.768  & 0.758 & 0.745 & 0.731  & 0.729 \\ \midrule
        $c_{20}$  & 0.782 & 0.780 & 0.777 & 0.768  & 0.757 & 0.743 & 0.734  & 0.728 \\ \midrule
        $c_{50}$  & 0.783 & 0.781 & 0.776 & 0.769  & 0.758 & 0.741 & 0.737  & 0.730 \\ \midrule
        $c_{200}$  & 0.784 & 0.781 & 0.777 & 0.769  & 0.760 & 0.744 & 0.744  & 0.738 \\ \midrule
         \bottomrule 
    \end{tabular}
    }
    % \end{adjustbox}
\vspace{-10pt}
\label{tab:aps}
\end{table}

\subsection{\response{Connection With Human Society}}
\response{Our main idea is similar to human society where people are not only focusing on the \textit{outcome justice}~\citep{oj} (e.g., fairness in the decision results) but pay increasing attention to the \textit{procedural justice}~\citep{pj} (e.g., fairness in the decision rationale). The regularization method to improve fairness can be deemed as achieving the \textit{outcome justice} directly. Our experiments/analysis show that \textit{procedural justice} might be easily violated in DNN models. We propose decision rationale alignment to further achieve the \textit{procedural justice} and improve fairness.}
% \subsection{\response{Various fairness metrics.}}
% \response{In our paper, we mainly focus on the metric demographic parity (DP) and the equalized odds (EO), both of which are introduced detailedly in the main paper. Our method is also applicable to other fairness metrics that quantify the expected difference between groups. For example, predictive parity focuses on whether the positive predictive value (PPV) is the same for both groups~\citep{Garg2020Fairness}. We should align the decision rationales for the data in both groups predicted as positive. 
% However, counterfactual fairness~\citep{Kusner2017Counterfactual} quantifies fairness from the perspective of an individual~\citep{Garg2020Fairness}, which is beyond our current framework. We will further explore it in the future.}

\subsection{\response{Combination With Data Augmentation}} 
\label{app:dataaug}
\response{The data augmentation and our decision rationale alignment are two independent ways to enhance fairness. From Fig. 3 (main paper), we can see that on the Credit dataset, FairReg($\Delta \text{DP}$, Aug) achieves better results than DRAlign under the DP metric. Intuitively, we can combine the two solutions straightforwardly. For example, we can replace the second term in Eq.(6) (main paper) (i.e., $L_\text{fair}$) with the data augmentation-embedded term (See \citep{fairmixup} for more details) and have a new formulation of Eq.(6) (main paper).}

\begin{align} 
\mathcal{L}= \text{E}_{(\mathbf{x},y) \sim P}(\mathcal{L}_{\text{cls}}(\text{F}(\mathbf{x}),y)) + \lambda \mathcal{L}_\text{aug}(\text{F}) + \beta \sum_{k=0}^Kd_k,
\end{align}

\response{We denote the above method for DP regularization as $\text{DRAlign}(\Delta \text{DP}, \text{Aug})$. We evaluate this version and compare it to the method without augmentation (i.e., $\text{DRAlign}(\Delta \text{DP})$ on the Credit dataset.  We see that: the fairness score (i.e., -DP) increases from -0.0169 to -0.0155 while the average precision (AP) also increases from 0.877 to 0.881, which further demonstrates the scalability of our method.}

\begin{wrapfigure}{r}{0.5\textwidth}
\includegraphics[width=1.0\linewidth]{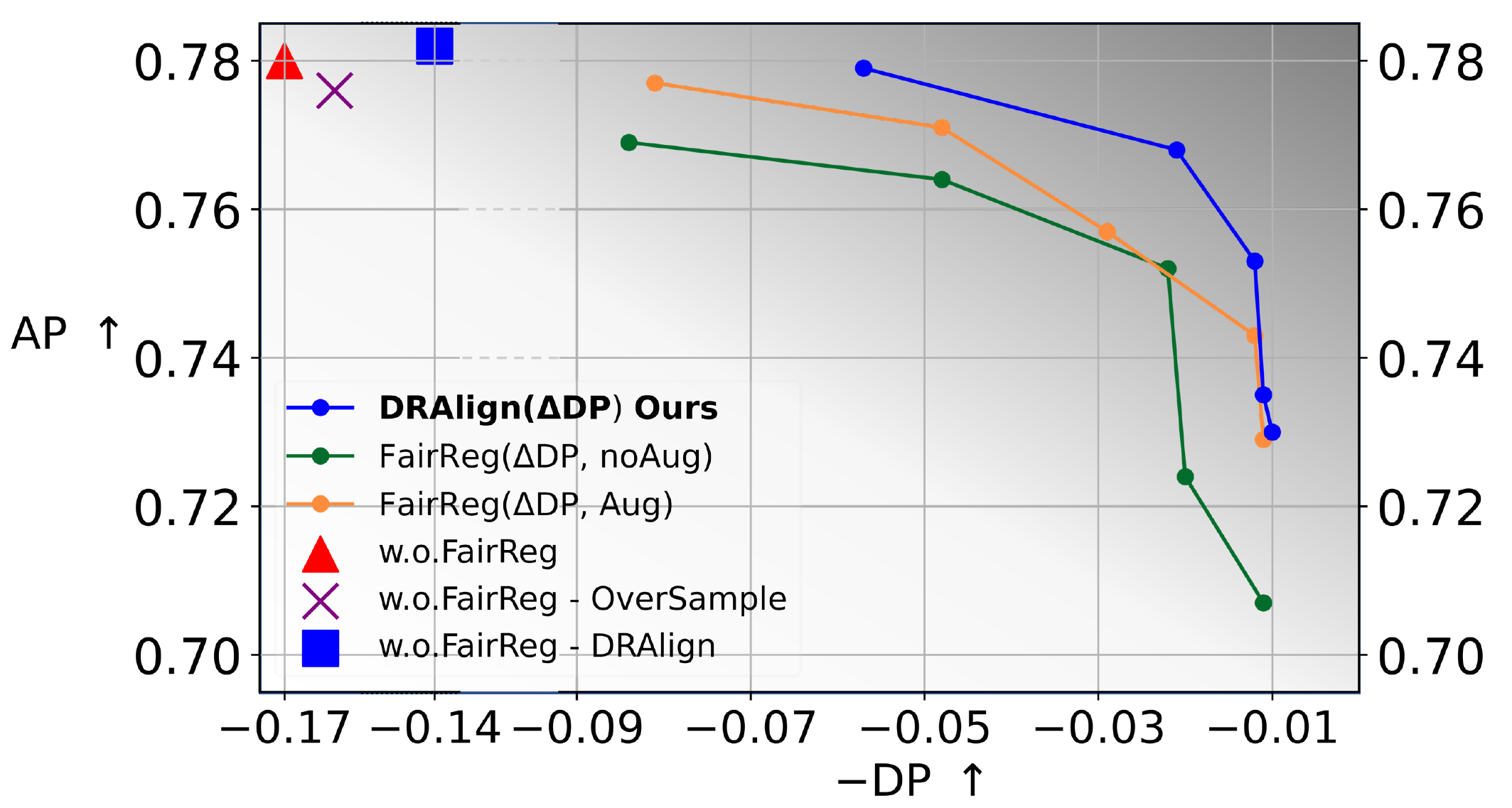} 
\caption{Accuracy and fairness comparison of five different methods on the Adult dataset. The hyperparameter $\lambda$ increases from 0.2 to 0.6 along the $-\text{DP}$ axis as it becomes larger.}
\label{fig:adult_DP2}
\end{wrapfigure}

\subsection{\response{Decision Rationale Alignment Without The Fairness Regularization.}}
\label{app:nofairreg}
% \begin{figure}
% \centering
% \includegraphics[width=0.5\linewidth]{pics/fig_adult_DP2.pdf}
% \caption{Accuracy and fairness comparison of five different methods on the Adult dataset. The hyperparameter $\lambda$ increases from 0.2 to 0.6 along the $-\text{DP}$ axis as it becomes larger.} 
% % \litl{data changed.}}
% \label{fig:adult_DP2}
% \vspace{-10pt}
% \end{figure}

\response{We find that the alignment itself could still slightly improve fairness when fairness regularization is removed. Specifically, we remove the $L_\text{fair}$ term in Eq.(6) (main paper) and retain the classification loss and the decision rationale alignment loss and compare the results of the two loss functions $L = L_{cls}$ and $L = L_{cls} + L_{DRA}$. We denote this version as $\text{w.o.FairReg-DRAlign}$. From Fig. \ref{fig:adult_DP2} we can see that: compared with the model only trained with the classification loss (i.e., $\text{w.o.FairReg}$, $\text{w.o.FairReg - Oversample}$), $\text{w.o.FairReg-DRAlign}$ increases the experimental results (AP, -DP) from (0.776, -0.16 ) to (0.781, -0.14). The results are consistent with our observation that our decision rationale alignment method could further improve fairness and demonstrate that decision rationale alignment is actually a favorable supplement for existing fairness regularization terms.}

\begin{table}[h] 
    \centering
    \caption{\response{T-test results for \text{FairReg}($\Delta \text{DP}$,noAug) and our DRAlign method on the Adult Dataset. }}
    % \begin{adjustbox}{max width=\textwidth,keepaspectratio}
    \resizebox{0.5\columnwidth}{!}{%
    \begin{tabular}{c|c|c|c|c}
    \toprule
        & $\lambda=0.2$ & $\lambda=0.3$ & $\lambda=0.4$ & $\lambda=0.5$ \\ \midrule
        P Value (Adult) & 1.19e-05 & 3.74e-05 & 0.011 & 0.027 \\ \midrule
         \bottomrule 
    \end{tabular}
    }
    % \end{adjustbox}
\label{tab:ttest1}
\end{table}

\begin{table}[h] 
    \centering
    \caption{\response{T-test results for \text{FairReg}($\Delta \text{DP}$,noAug) and our DRAlign method on the CelebA dataset. }}
    % \begin{adjustbox}{max width=\textwidth,keepaspectratio}
    \resizebox{0.5\columnwidth}{!}{%
    \begin{tabular}{c|c|c|c|c}
    \toprule
        & $\lambda=0.2$ & $\lambda=0.3$ & $\lambda=0.4$ & $\lambda=0.5$ \\ \midrule
        P Value (CelebA) & 0.017 & 0.018 & 0.003 & 0.002 \\ \midrule
         \bottomrule 
    \end{tabular}
    }
    % \end{adjustbox}
\label{tab:ttest2}
\end{table}

\begin{table}[h] 
    \centering
    \caption{\response{Comparison between DRAlign(ours) and FairReg(*, noAug) under the EOP metric and the PP metric.}}
    % \begin{adjustbox}{max width=\textwidth,keepaspectratio}
    \resizebox{0.8\columnwidth}{!}{%
    \begin{tabular}{c|c|c|c|c|c|c|c}
    \toprule
        &\multicolumn{3}{c|}{Equality of Opportunity}  &  &\multicolumn{3}{c}{Predictive Parity}\\ \midrule
        % & $\lambda = 0.03$ & $\lambda = 0.1$   & $\lambda = 0.5$  & &  $\lambda = 0.03$  & $\lambda = 0.03$   & $\lambda = 0.03$ \\ \midrule
        $AP_{DRAlign}$ & \textbf{0.7787} & \textbf{0.7737}  & \textbf{0.7597} &$AP_{DRAlign}$ &  \textbf{0.7854} & \textbf{0.7852}  & 0.7839\\ \midrule
        $EOP_{DRAlign}$ & \textbf{0.9494} & \textbf{0.9510} & \textbf{0.9692} &$PP_{DRAlign}$ &  \textbf{0.0219} & \textbf{0.0186} & \textbf{0.0175} \\ \midrule
        $AP_{FairReg(\Delta EOP, noAug)}$ & 0.7769 & 0.7695 & 0.7552 & $AP_{FairReg(\Delta PP, noAug)}$ &  0.7852 & 0.7845 &  \textbf{0.7840}\\ \midrule
        $EOP_{FairReg(\Delta EOP, noAug)}$ & 0.9411 & 0.9456 & 0.9606 & $PP_{FairReg(\Delta PP, noAug)}$&  0.0276 &  0.0246 & 0.0205 \\ \midrule
         \bottomrule 
    \end{tabular}
    }
    % \end{adjustbox}
\label{tab:eop}
\end{table}

\begin{table}[htb]
\caption{Experimental results on the COMPAS dataset.} 
\begin{tabular}{c|cc|cc|c|c|c|c|c}

\hline
Method            & \multicolumn{2}{c|}{}                             & \multicolumn{2}{c|}{}                              & Method            &              &               &               &               \\ \hline
Van               & \multicolumn{1}{c|}{AP}           & 0.649         & \multicolumn{1}{c|}{}              &               & Van               & AP           & 0.643         &               &               \\ \hline
  & \multicolumn{1}{c|}{-DP}          & -0.245        & \multicolumn{1}{c|}{}              &               &                   & -EO          & -0.442        &               &               \\ \cline{1-10} 
  Oversampling   & \multicolumn{1}{c|}{AP}           & 0.636         & \multicolumn{1}{c|}{}              &               & Oversampling      & AP           & 0.65          &               &               \\ \cline{1-10} 
                  & \multicolumn{1}{c|}{-DP}          & -0.189        & \multicolumn{1}{c|}{}              &               &                   & -EO          & -0.102        &               &               \\ \hline
FairReg(DP,noAug) & \multicolumn{1}{c|}{\textbf{lam}} & \textbf{0.05} & \multicolumn{1}{c|}{\textbf{0.1}}  & \textbf{0.3}  & FariReg(EO,noAug) & \textbf{lam} & \textbf{0.02} & \textbf{0.03} & \textbf{0.05} \\ \hline
                  & \multicolumn{1}{c|}{AP}           & 0.637         & \multicolumn{1}{c|}{0.635}         & 0.634         &                   & AP           & 0.647         & 0.646         & 0.646         \\ \hline
                  & \multicolumn{1}{c|}{-DP}          & -0.117        & \multicolumn{1}{c|}{-0.055}        & -0.015        &                   & -EO          & -0.045        & -0.044        & -0.034        \\ \hline
FairReg(DP,Aug)   & \multicolumn{1}{c|}{\textbf{lam}} & \textbf{0.02} & \multicolumn{1}{c|}{\textbf{0.04}} & \textbf{0.05} & FairReg(EO, Aug)  & \textbf{lam} & \textbf{0.02} & \textbf{0.03} & \textbf{0.05} \\ \hline
                  & \multicolumn{1}{c|}{AP}           & 0.638         & \multicolumn{1}{c|}{0.637}         & 0.636         &                   & AP           & 0.648         & 0.647         & 0.644         \\ \hline
                  & \multicolumn{1}{c|}{-EO}          & -0.145        & \multicolumn{1}{c|}{-0.122}        & -0.105        &                   & -EO          & 0.07          & 0.06          & 0.032         \\ \hline
DRAlign           & \multicolumn{1}{c|}{\textbf{lam}} & \textbf{0.05} & \multicolumn{1}{c|}{\textbf{0.1}}  & \textbf{0.3}  & DRAlign           & \textbf{lam} & \textbf{0.02} & \textbf{0.03} & \textbf{0.05} \\ \hline
                  & \multicolumn{1}{c|}{AP}           & 0.639         & \multicolumn{1}{c|}{0.637}         & 0.635         &                   & AP           & 0.649         & 0.647         & 0.646         \\ \hline
                  & \multicolumn{1}{c|}{-DP}          & -0.104        & \multicolumn{1}{c|}{-0.033}        & -0.008        &                   & -EO          & -0.044        & -0.039        & -0.032        \\ \hline
\end{tabular}
\label{tab:compas_result}
\end{table}

\subsection{\response{The T-Test Results to Show The Significant Fairness Improvement}}
\label{app:ttest}
% \begin{figure*}
%     \centering
%     \includegraphics[width=1\linewidth]{pics/error_bar.pdf}
%     \caption{Accuracy and fairness comparison with error bar.}
%     \label{fig:errorbar}
%     \vspace{-0.1in}
% \end{figure*}

% \response{Here, we only show the error bar of our experimental results in Fig. \ref{fig:errorbar} on FairReg($\Delta$DP, noAug) and DRAlign for better observation. It should be noted that the x-coordinate -DP and y-coordinate AP are both changing with the random seed. We here plot a rectangular region for the error bar of each data point. Moreover, we mark a point "A" in the figure of the Adult dataset under -DP metric. Although we plot a rectangular region for the error bar, it does not mean that point A (-0.073, 0.777) can be reached by FairReg($\Delta DP$, noAug). It just means the -DP value and the AP value of FairReg($\Delta DP$, noAug) could arrive at -0.073 and 0.777 separately with different random seeds.}

\response{We showcase our results are statistically significant via t-test (regarding \text{FairReg}($\Delta \text{DP}$,noAug) and DRAlign under the DP metric on the Adult dataset and the CelebA dataset in Table \ref{tab:ttest1} and Table \ref{tab:ttest2}. We can see that under all parameter settings, the p-values < 0.05. This means that our results are statistically significant. Note that, our method not only improves the model performance in fairness but also improves the model accuracy. Taking both AP values and fairness performance into consideration, our DRAlign outperforms other methods saliently.}

\subsection{\response{Experiments on More Fairness Measures}}
\label{app:eop_pp}
we further evaluate and compare our method with FairReg methods on the third popular metric Equality of Opportunity (EOP), and Predictive Parity (PP). Specifically, we adopt the EOP definition in \cite{wang2022mitigating}, and the PP definition in \cite{Garg2020Fairness}.

$$EOP = TPR_{a=0} / TPR_{a=1} = P(\hat{y} =1| a=0, y=1) /  P(\hat{y}=1|a=1,y=1)$$

$$PP = | p(y = 1|a = 0, \hat{y} = 1) - p(y = 1|a = 1, \hat{y} = 1) |.$$

Under the above definition, EOP close to 1 and PP close to 0 indicate fair classification results. We carefully modify the FairReg method for the EOP and PP metrics.

$$
L_{fair, EOP} = \Delta \text{EOP}(F) =  E_{\mathbf{x} \sim P_0^1}(F(\mathbf{x})) - E_{\mathbf{x} \sim P_1^1}(F(\mathbf{x})),
$$

$$
L_{fair, PP} =  \Delta \text{PP}(F) =  \frac{ E_{\mathbf{x} \sim P_0^0}(F(\mathbf{x})) * N_0^0) + ( E_{\mathbf{x} \sim P_0^1}(F(\mathbf{x})) * N_0^1)}{N_0^0+N_0^1}   - \frac{ E_{\mathbf{x} \sim P_1^0}(F(\mathbf{x})) * N_1^0) + ( E_{\mathbf{x} \sim P_1^1}(F(\mathbf{x})) * N_1^1)}{N_1^0+N_1^1} ,
$$

$N_0^0$, $N_0^1$, $N_1^0$, $N_1^1$ are the sample numbers of subgroups $D_{00}$, $D_{01}$, $D_{10}$, and $D_{11}$, which satisfy the following attribute and category conditions:  {a=0,y=0}, {a=0,y=1},{a=1,y=0},{a=1,y=1} in the batch of data. The sampling methods of FairReg($\Delta$ \text{EOP}, noAug) and FairReg($\Delta$ \text{PP}, noAug) follow those of FairReg($\Delta$ \text{EO}, noAug). For the EOP metric, we align the decision rationales between subgroups {a=0,y=1} and {a=1,y=1}. For the PP metric, we align the decision rationales of ($D_{00}$, $D_{10}$) and the decision rationales of ($D_{01}$,$D_{11}$).

We showcase the experimental results here (the report averages over 10 times). The models evaluated on the EOP metrics are trained for 20 epochs. $\lambda$ is set as in range \{0.03,0.1,0.5\} and $\beta$ is set as $\lambda / 10$. The models evaluated on the EOP metrics are trained for 5 epochs. $\lambda$ is set as in range \{0.07,0.09,0.1\}. From the Table \ref{tab:eop}, we can see that our method DRAlign consistently improves the fairness performance under the EOP and PP metrics,  that is, our method could be extended to EOP and PP.

\subsection{Experiments on the COMPAS dataset}

We further evaluate another widely-used dataset, i.e., COMPAS \cite{compas}. We train all the models in 5 epochs. The learning rate is set as 0.001. For our method DRAlign, 
 is set as $\beta/10$. From Table \ref{tab:compas_result}, we see that: our method consistently outperforms all baseline methods, which is in line with the conclusion of our paper. For instance, when we set $\lambda$ as 0.1, DRAlign($\Delta$DP) achieves higher AP (0.637) and -DP (-0.033) scores than the baseline methods. We will include the results in our paper.

\subsection{Independent assumption across neurons}
\label{sec:assumptions}
The assumption of neuron independence is used to estimate the influence/importance of a group of neurons by calculating the influence/importance of each individual neuron. In our work, the essential objective of decision rationale alignment is to ensure that eliminating any random combination of neurons has the same influence on various subgroups. However, evaluating the importance of all potential neuron combinations is impractical and computationally challenging due to the vast search space for possible combinations of neurons. Therefore, we assume that neurons are independent, which enables us to estimate the influence of any neuron combination by summing the individual impacts of each neuron in the combination. This assumption can be viewed as a "greedy" approximation strategy to assess the influence of arbitrary neuron combinations \cite{molchanov2016pruning,molchanov2019importance}. Note that the independence assumption doesn’t limit our method to networks with all independent neurons. Instead, it is an approximation to align the influence of any neuron combination in regular networks.

Note that, the neuron independence assumption has been widely used in previous works \cite{molchanov2016pruning,molchanov2019importance} to achieve the approximation to evaluate the influence of a group of neurons.

\subsection{Comparing pre-processing and post-processing techniques}

We further add experiments to compare our method with a pre-processing method which adjusts the inputs to be uncorrelated with the sensitive attribute in each iteration. This method achieves a -DP value of -0.086 and an AP value of 0.761. We test the parity score of the trained model and find it to be 0.43, which is smaller than the parity score of 0.624 obtained through standard training. This observation is consistent with our conclusion drawn using the FairReg(*,noAug) method and shows that the pre-processing method could also implicitly lead to lower neuron/parameter parity scores.

However, it should be noted that post-processing techniques only modify the outputs of the trained model and do not alter the model itself. Therefore, the post-processing methods could be regarded as aligning the neuron/parameter of the last layer while neglecting the alignment of the middle neurons/parameters. Thus, with these observations, we can regard the misalignment as a fundamental reason for unfairness. 

\end{document}